\documentclass[lettersize,journal]{IEEEtran}
\usepackage{amsmath,amsfonts}
\usepackage{algorithmic}
\usepackage{algorithm}
\usepackage{array}
\usepackage[caption=false,font=normalsize,labelfont=sf,textfont=sf]{subfig}
\usepackage{textcomp}
\usepackage{stfloats}
\usepackage{url}
\usepackage{verbatim}
\usepackage{graphicx}
\usepackage{cite}
\usepackage{booktabs}
\usepackage{multirow}
\usepackage{pdflscape}
\usepackage{adjustbox}
\usepackage{bbding}
\usepackage{pifont}
\usepackage{xcolor}
\usepackage{graphicx} 
\usepackage[normalem]{ulem}
\usepackage{comment}
\usepackage{amssymb}
\begin{document}


\title{Enhancing Knowledge Transfer in Hyperspectral Image Classification via Cross-scene Knowledge Integration}

\author{Lu Huo, \thanks{Lu Huo is with School of Electrical and Data Engineering, Faculty of Engineering and IT, University of Technology Sydney, Sydney, 2007, NSW, Australia} Wenjian Huang, \thanks{Wenjian Huang is with Research Institute of Trustworthy Autonomous Systems and Department of Computer Science and Engineering, Southern University of Science and Technology, Shenzhen 518055, China} Jianguo Zhang, \thanks{Jianguo Zhang is with Research Institute of Trustworthy Autonomous Systems and Department of Computer Science and Engineering, Southern University of Science and Technology, Shenzhen 518055, China} Min Xu, \thanks{Min Xu is with School of Electrical and Data Engineering, Faculty of Engineering and IT, University of Technology Sydney, Sydney, 2007, NSW, Australia} Haimin Zhang \thanks{Haimin Zhang is with School of Electrical and Data Engineering, Faculty of Engineering and IT, University of Technology Sydney, Sydney, 2007, NSW, Australia}
\thanks{Corresponding authors: Jianguo Zhang and Min Xu}
}
\markboth{Journal of \LaTeX\ Class Files,~Vol.~14, No.~8, August~2021}%
{Shell \MakeLowercase{\textit{et al.}}: A Sample Article Using IEEEtran.cls for IEEE Journals}


\maketitle

\begin{abstract}
In recent years, hyperspectral imaging (HSI) has emerged as a critical tool in remote sensing, offering detailed spectral information for environmental monitoring, agriculture, and mineralogy. Despite its potential, the effective analysis and application of knowledge transfer using HSI data are hindered by critical challenges: \emph{spectral discrepancies} caused by different HSI sensors and \emph{mismatched semantics} resulting from a lack of category correspondence. These challenges pose significant obstacles to knowledge sharing across scenes, particularly limiting model performance when the target task has a small amount of training data. Some preliminary works have utilized a shared data portion for partial knowledge transfer, resulting in insufficient knowledge sharing and incomplete utilization of information for the target task. 
In addition, existing settings for HSI knowledge transfer primarily focus on homogeneous contexts within the same dataset or heterogeneous settings with co-occurrence categories.
These limitations restrict their applicability.
To navigate these challenges, we introduce a novel methodology, Cross-scene Knowledge Integration (CKI). 
First, we propose the Alignment of Spectral Characteristics (ASC) to address \emph{spectral discrepancies} by projecting the data from two distinct domains into a domain-agnostic space. Next, we introduce Cross-scene Knowledge Sharing Preference (CKSP) to resolve \emph{mismatched semantics} by employing a Source Similarity Mechanism (SSM) to determine the importance of source samples during knowledge sharing. Finally, to address the deficiency in utilizing target-private information within shared knowledge, we implement Complementary Information Integration (CII) using complementary extraction and distillation integration modules. This ensures the completeness of the target information.
Extensive testing across diverse knowledge transfer settings reveals that the proposed CKI not only maintains stable performance but also significantly outperforms existing knowledge transfer methods in the HSI field, achieving state-of-the-art results. 

\end{abstract}


\section{Introduction}
Hyperspectral Imaging (HSI) is a technology used in remote sensing and imaging spectroscopy. It involves capturing and processing electromagnetic radiation based on the molecular composition and texture of different objects \cite{ghamisi2017advances}. 
From a spectral response perspective, the concept of the spectrum is key to distinguishing different surface covers because of their unique spectral characteristics. In any given dataset, different surface cover classes are characterized by their own unique families of spectral responses \cite{zhang2016simultaneous,Landgrebe2002HyperspectralID}. However, in real remote sensing applications, the need often arises to classify a particular HSI scene (referred to as the target scene) that has very few labeled samples, a situation typically resulting from certain natural limitations and high labor costs of labeling \cite{ye2017dictionary}. 


To overcome the challenge of limited target data, existing methods utilize information from a source scene that possesses sufficient labeled samples. 
Knowledge transfer methods in HSI primarily focus on homogeneous settings, transferring knowledge from the source area to the target area within the same dataset \cite{wang2020hyperspectral,ye2017dictionary,yu2021unsupervised,deng2018active}, and on heterogeneous settings, where co-occurrence classes from source and target scenes are matched first \cite{ning2023contrastive,ye2022learning,chen2020semisupervised,ye2024adaptive}. 
These two settings help avoid inconsistent categories, while the source-private knowledge could adversely affect the performance of the target model\cite{day2017survey,zhang2022survey}.
In cross-scene HSI datasets, the lack of standardized criteria for HSI data acquisition leads to challenges in knowledge transfer due to spectral shifts and inconsistencies in the number of spectral bands \cite{Landgrebe2002HyperspectralID,huang2023cross}. Accordingly, after matching the consistent categories, current HSI knowledge transfer methods generally use domain adaptation (DA) techniques to align spectral shifts and/or spectral distributions \cite{yu2021unsupervised, fang2022confident, peng2022domain,huang2024adversarial,li2023deep,huang2023cross,li2023supervised,qu2023feature,qu2024cycle,yu2024hyperspectral,huang2024adversarial}. However, real HSI knowledge transfer scenarios are more complex, with differences in both feature and class spaces \cite{peng2022domain}.
Subsequently, the cross-scene inconsistent-categories setting has been proposed, characterized by discrepancies in spectral domains and unmatched categories \cite{9356245}. Combining domain alignment to handle varying spectral domains with few-shot learning for inconsistent categories is a common method in this setting \cite{ye2023adaptive,xu2023graph,li2024scformer,zhang2023cross,wang2024dual,10299721}. However, this approach has an inherent constraint: the number of categories in the source domain must exceed those in the target scene \cite{9356245}, which limits the application scope of HSI knowledge transfer.


Overall, these settings fail to meet the demands of real-world remote sensing classification for HSI, as real-world scenarios typically involve more complex cross-scene datasets that face \textit{\textbf{spectral discrepancies}} caused by different HSI sensors and uncertain, \textit{\textbf{semantically mismatched categories}} due to the lack of direct label correspondence between source and target scenes \cite{peng2022domain}. 
As shown in Table \ref{example_table} and Fig. \ref{examplew_picture}, the differences in data acquisition and categories coexist. Three benchmark datasets listed in Table \ref{example_table} were captured by distinct HSI sensors, each with different ranges of wavelengths and spectral bands, indicating significant spectral discrepancies among various sensors.
In addition, the category space is semantically mismatched, meaning that some samples from source categories may correspond to parts of target categories labeled from different semantic perspectives. 
As shown in Fig. \ref{examplew_picture}, there is no straightforward, one-to-one correspondence between the semantically related labels in the source (``Buildings-Grass-Trees-Drives", ``Corn", and ``Woods") and the target (``Road", ``Residential", ``Commercial", ``Parking Lot", ``Grass", ``Trees") scenes. 

Moreover, in addition to the semantically mismatched categories, non-overlapping semantics between the source and target tasks also exist. As shown in Fig. \ref{examplew_picture}, both the source and target scenes contain outlier classes situated in non-overlapping spaces (specifically, ``Corn" in the source and ``Parking Lot" in the target). Existing studies primarily focus on the shared portion between the source and target scenes, leading to a \textit{\textbf{deficiency in utilizing target-private information within shared knowledge}}, which can result in suboptimal performance for the target task \cite{liu2023category, ding2016incomplete, li2020domain, gong2020cross}.
As a result, deducing the complete information for the target scene solely from the overlapping sections between the source and target is a challenge. 

In a nutshell, given the presence of spectral discrepancies and semantic mismatches in cross-scene datasets, designing an effective knowledge transfer method that can efficiently utilize the existing source dataset while also fully leveraging the discriminative information of the target task remains a significant challenge. Specifically, in this context, we encounter three challenges:

\begin{figure}[!t]
\includegraphics[width=\columnwidth]{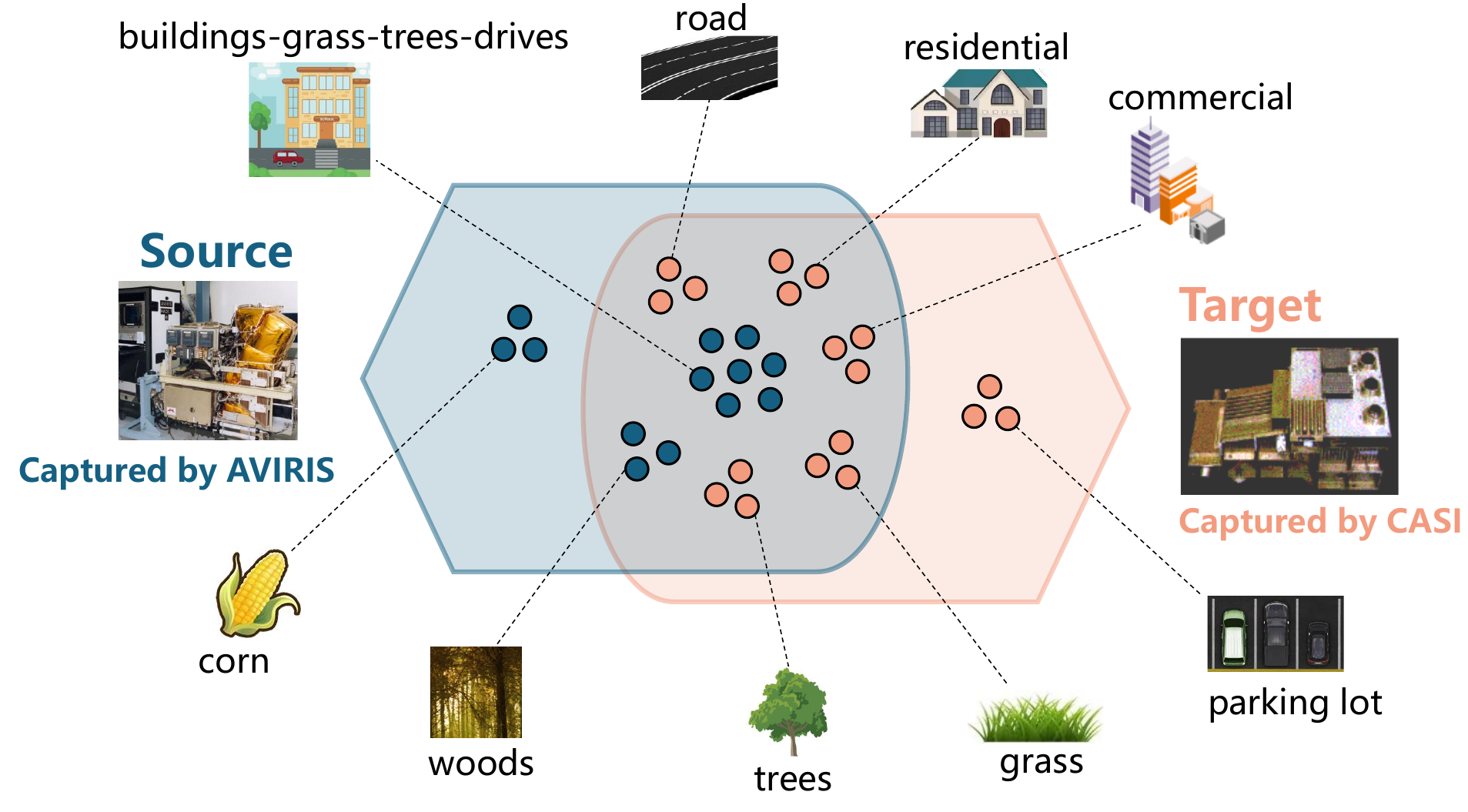}
\caption{The general scenario studied in this paper, where source and target scenes are characterized by disparate spectral features attributable to distinct HSI sensors such as AVIRIS and CASI. Furthermore, there exists a mismatch in the label spaces between the label spaces of the source (encompassing categories including ``Buildings-Grass-Trees-Drives", ``Corn", and ``Woods") and the target scenes (``Road", ``Residential", ``Commercial", ``Parking Lot", ``Grass", ``Trees"). The intersection of these label spaces contains coarse-grained categories (e.g., ``Buildings-Grass-Trees-Drives” and ``Woods") and fine-grained categories (e.g., ``Residential”, ``Commercial", and ``Grass”). In contrast, outlier classes—--particularly ``Corn" from the source and ``Parking Lot" from the target scenes---lie outside this overlap.}
\label{examplew_picture}
\end{figure}


\begin{enumerate}
\item{\emph{Spectral discrepancies} exist between the distinct wavelength ranges of various HSI sensors, resulting in significant domain gaps between the source and target domains, and hindering cross-scene knowledge transfer. When the characteristics of the source and target domains diverge, models trained on the source domain often fail to generalize effectively to the target.
}
\item{\emph{Semantic mismatches} arising from category distinctions between the source and target scenes pose significant challenges for classification tasks in HSI. These distinctions can lead to poor generalization in target classes when tuning models initially trained on the source scene.} 
\item{Due to inherent discrepancies in cross-scene datasets, focusing solely on knowledge transfer from shared portion can result in \emph{a deficiency in target-private information utilization}. The challenge of fully extracting the discrimination information for the target task remains to be solved when using the shared information of the source and target data.}
\end{enumerate}

To solve these challenges, we propose a Cross-scene Knowledge Integration (CKI) method that addresses a generalized scenario for cross-scene knowledge transfer, where spectral discrepancies and semantically mismatched categories coexist.
Specifically, the Alignment of Spectral Characteristics (ASC) module is proposed to address \emph{spectral discrepancies} by using adversarial learning to create a domain-agnostic space that aligns distinct spectral distributions. Additionally, the Cross-scene Knowledge Sharing Preference (CKSP) module is introduced to resolve \emph{mismatched semantics} by identifying the shared semantics between the source and target, facilitating knowledge transfer through sample-level semantic correspondence by introducing the Source Similarity Mechanism (SSM). Finally, the Complementary Information Integration (CII) module is proposed, which identifies target-private information to supplement the shared knowledge portion through distillation integration, addressing the \emph{deficiency in target-private information utilization} and ensuring the completeness of the target information.

The main contributions of this paper can be summarized as follows:
\begin{enumerate}
\item{We introduce the ASC component to address spectral discrepancies across various HSI sensors. This ensures the effective knowledge transfer in a domain-agnostic space.}
\item{We propose the CKSP component to address the lack of category correspondence between the source and target. This overcomes traditional barriers caused by mismatched semantics.}
\item{
 We develop the CII component to explicitly incorporate target-private information, alleviating the issue of focusing solely on partially shared knowledge. This enhances the quality and completeness of the target information.}
\item{In the general context of HSI cross-scene knowledge transfer, the proposed CKI significantly improves the accuracy of the target task, achieving state-of-the-art performance compared to traditional approaches.}
\end{enumerate}

The rest of this article is structured as follows: Section II provides a summary of knowledge transfer methods and relevant applications in HSI. The technical challenges and the proposed CKI are described in detail in Section III. The experimental results are presented and analyzed in Section IV. Finally, the conclusions are outlined in Section V.


\begin{table}[]
\centering
\caption{The data acquisition and class information of each scene}
\begin{tabular}{c|c|c|c|c}
\hline
\bottomrule
 Dataset & Sensor & Wavelength Range & Class & Bands \\ \hline
Indian Pines & AVIRIS & 0.40-2.50 µm & 16 & 200 \\ \hline
Pavia University & ROSIS-3 & 0.43-0.86 µm & 9 & 103 \\ \hline
Houston 2013 & CASI-1500 & 0.38-1.05 µm & 15 & 144 \\ \hline
\bottomrule
\end{tabular}
\label{example_table}
\end{table}

\section{Related Work}

\subsection{Knowledge Transfer}
Knowledge transfer in the context of machine learning and artificial intelligence refers to the process by which knowledge learned in one task or domain is applied or adapted to improve learning in a related but different task or domain \cite{lu2015transfer,pan2009survey,yang2020transfer,zhou2022domain,wang2022generalizing}. 
This concept can generally be categorized into the following components:
\subsubsection{Domain Adaptation (DA)}
In the context of DA \cite{ben2006analysis}, there are several paradigms designed to address the challenges of transferring knowledge from a labeled source domain to an unlabeled \cite{ganin2015unsupervised} or partially labeled \cite{pan2010domain} target domain, where the data may differ but remain related \cite{csurka2017domain}. These paradigms are broadly categorized into closed-set \cite{Tzeng_2017_CVPR}, partial \cite{cao2018partial}, and open-set domain adaptation \cite{panareda2017open}, each tailored to specific scenarios of domain shift and label space differences \cite{you2019universal}. 
However, existing studies typically focus on homogeneous inputs from the source and target domains that share the same format, such as RGB images \cite{csurka2017domain} and NLP data \cite{ramponi2020neural}. In many existing cases, the source and target categories often exhibit a clear correspondence or are directly considered orthogonal. There is a deficiency in explicitly modeling the semantic relations between the source and target labels when they are semantically related but lack direct correspondence.

\subsubsection{Heterogeneous Transfer Learning (HTL)}
Heterogeneous Transfer Learning involves source and target domains with differing feature spaces and may also include challenges such as varying data distributions and label spaces \cite{day2017survey}. 
In this scenario, the source and target domains may share no common features or labels, and the dimensions of their input feature spaces may also differ \cite{zhuang2020comprehensive,bao2023survey}.
When the target domain lacks adequate labeled data, semi-supervised settings are often used to address this challenge to a certain extent \cite{li2013learning,zhu2011heterogeneous, fang2022semi,feng2022semi}.
However, these semi-supervised settings require that the categories of the source and target domains must be perfectly consistent \cite{li2024scformer}.
The primary challenge in a heterogeneous setting is the risk of negative transfer, where misleading or irrelevant knowledge from the source domain is applied to the target domain \cite{bao2023survey}.
Moreover, developing algorithms that can effectively align and map the inherent differences in feature spaces from one modality to another, while retaining essential and distinctive features and avoiding negative transfer, is crucial for the target task \cite{bao2023survey}.

\subsubsection{Parameter Efficient Fine-Tuning (PEFT)}
From a parameter-efficient perspective of knowledge transfer, strategies such as fine-tuning a small number of parameters \cite{he2021towards}, employing adapters \cite{houlsby2019parameter}, utilizing prompting techniques \cite{li2021prefix}, and applying low-rank matrix factorization \cite{hu2021lora} for approximate parameter updates stand out. These approaches enable the effective adaptation and customization of pre-trained models to new tasks or domains. Furthermore, these approaches optimize the reuse of learned features, thereby reducing the need for extensive data collection and training from scratch \cite{chen2022revisiting}. 
However, these strategies generally require large-scale pre-trained datasets and substantial computational resources to develop satisfactorily generalizable foundation models. Moreover, their effectiveness is often limited by the inherent challenges associated with aligning the pre-trained features with the specific requirements of the target task \cite{zhang2022survey,rosenstein2005transfer,pan2009survey}. 

\subsubsection{Knowledge Distillation (KD)}

Knowledge Distillation is an approach initially designed to transfer knowledge from a comprehensive, often complex model (referred to as the ``teacher") to a simpler, more compact model (known as the ``student") via soft labels or logits for the same task \cite{hinton2015distilling, buciluǎ2006model, gou2021knowledge}. 
The teacher model is commonly designed for source data, whereas the student model is tailored for target data.
Various distillation strategies, including online distillation and self-distillation \cite{zhu2018knowledge, li2022shadow, li2022distilling}, have been developed to enhance the performance of the student model, especially in the absence of a large-capacity, high-performance teacher model \cite{gou2021knowledge}. Recent studies have demonstrated the potential of Distillation from Weak Teacher (DWT), which improves the performance of a larger student model by transferring knowledge from a weaker, smaller teacher model \cite{lee2023study}. 
However, there remain gaps in exploring how to supplement knowledge through distillation. In the context of cross-scene HSI knowledge transfer investigated in this study, the differences between the source task and the target task, reflected in the heterogeneity of data domains and partially overlapping semantic label spaces, may cause the transfer model to focus only on partially shared discriminative features. As a result, the extraction of discriminative features relevant to the target task can be incomplete.





\subsection{Knowledge Transfer in HSI}
Existing knowledge transfer methods in the HSI field can be broadly classified into the following categories:
\subsubsection{Homogeneous Knowledge Transfer}
Homogeneous setting incorporates information from the source area into the target area within the same dataset.
The main challenge of homogeneous knowledge transfer is spectral shift, where the spectral distributions of the same class may deviate significantly across different areas in the same dataset \cite{ye2017dictionary}.

To mitigate this challenge, Active Learning (AL) and DA techniques based on multiple kernels \cite{deng2018active}, dictionary learning combined with multitask learning \cite{ye2017dictionary} and Maximum Mean Discrepancy (MMD) with manifold regularization \cite{wang2020hyperspectral} have been developed to compensate for spectral shifts between two separate HSI areas. 
Nevertheless, the homogeneous setting is an ideal case in HSI knowledge transfer. In real situations, knowledge transfer in HSI can be more complex, as the data acquisition process and/or objects may differ between the source and target HSI data.

\subsubsection{Heterogeneous Knowledge Transfer}
A heterogeneous setting involves cross-scene data from distinct datasets, enabling the transfer of knowledge regarding similar or identical materials from the source to the target scene. 
Cross-scene data may be captured from different HSI devices, leading to inconsistent spectral channels and/or different objects across scenes, which complicates knowledge transfer between datasets \cite{Landgrebe2002HyperspectralID, ye2017dictionary}.
Overall, HSI cross-scene knowledge transfer comprises two situations: co-occurrence and inconsistent categories.

Co-occurrence indicates the shared classes are selected first before matching the spectral distributions.
Researchers initially chose shared classes across different HSI scenes and manually aligned the wavelength ranges for different HSI sensors or restricted the use to the same HSI sensors \cite{peng2022domain,fang2022confident,zhang2021topological}.
Confident Learning-based Domain Adaptation \cite{fang2022confident}, 
nonlocal topological relationships coupled with Semantic Information \cite{zhang2021topological}, and instance-to-instance contrastive learning \cite{ning2023contrastive} have been proposed to decrease spectral domain discrepancy.
Subsequently, researchers have focused on matching the common categories first before performing knowledge transfer in heterogeneous cross-scene HSI datasets.
Adaptive Graph Modeling coupled with Self-Training \cite{ye2024adaptive}, Multiscale Convolutional Sparse Decomposition (MCSD) and Structure-Preserved Distribution Alignment (SPDA) \cite{zhong2022heterogeneous}, cross-domain variational autoencoder combined with graph regularization \cite{ye2022learning}, and Attention Block with Cross-Domain Loss \cite{wang2024dual} have been introduced to mitigate domain shift.



Inconsistent categories relaxes the category restrictions, enabling HSI cross-scene knowledge transfer to occur in a general scenario. 
To minimize dependency on paired data and address inconsistent categories, multiple related HSI datasets are used as source data to improve generalization performance through Multi-Task Learning (MTL) \cite{lee2022exploring, gao2023learning}.
Later, Few-Shot Learning (FSL) has been used to address the situation of inconsistent categories \cite{liu2018deep,10299721,bai2022few,huang2023hyperspectral,ye2024cross,li2024cross, qin2024few,hu2023cross}.
An increasing number of researchers combine DA with FSL to align domain shifts and solve inconsistent categories situation \cite{9356245,wang2022spatial,ye2023adaptive,xu2023graph,ding2024glgat,zhang2023cross,wang2024dual}.

Although existing heterogeneous knowledge transfer methods have addressed some issues in cross-scene HSI, they retain an inherent limitation when only a single source HSI dataset is considered, particularly in FSL. The source typically has more categories than the target \cite{9356245,9356245,wang2022spatial,ye2023adaptive,xu2023graph,ding2024glgat,zhang2023cross,wang2024dual}, restricting the applicability of HSI Heterogeneous Knowledge Transfer (HKT) scenarios.

\section{Cross-scene Knowledge Integration}
\subsection{Problem Setting}
Let's consider two scenes: the source data and the target data with device variability and category mismatches.
Source domain $D_s$ comprise $N_s$ labeled samples,
$D_s = \{(x^s_{i}, y^s_{i})\}_{i=1}^{N_s}$, where $x^s_{i} \in \mathbb{R}^{d_s}$ represent the HSI input vector of the $i$-th sample in the source domain with dimensionality $d_s$ and $y^s_{i} \in C_s$ is its corresponding label from the source label set $C_s$. In contrast, target domain $D_t$ contains $N_t$ labeled samples ($N_t\ll N_s$),
$D_t = \{(x^t_{j}, y^t_{j})\}_{j=1}^{N_t}$, where $x^t_{j} \in \mathbb{R}^{d_t}$ is the HSI input vector of the $j$-th sample in the target domain with dimensionality $d_t$ (usually $d_s\neq d_t$) and $y^t_{j} \in C_t$ represent its corresponding label from the target lable set $C_t$ (usually $C_s\neq C_t$).

\subsection{Technical Challenges}
Different HSI sensors may have varying spectral resolutions, sensitivities, and calibration standards due to differences in sensor characteristics, acquisition conditions, and pre-processing methods. These spectral discrepancies can significantly alter the captured spectral signatures of the same material, leading to variations in the data collected by each sensor. The resulting spectral variability appears as differences in spectral distributions or features, represented mathematically by $p(x^s_{i})$ for the source domain and $q(x^t_{j})$ for the target domain. Even in the presence of spectral discrepancies induced by distinct HSI sensors, mismatched semantics due to a lack of category correspondence remain in our setting. Unlike traditional hyperspectral imaging (HSI) domain adaptation, which typically assumes that the source and target domains share identical label spaces or are analyzed orthogonally in the semantic space for non-identical settings, the proposed Cross-scene Knowledge Integration (CKI) instead allows for partially overlapping label spaces semantically.

The hard matching in the label space and the feature domain used by matching-based strategies, or the blind alignment in the embedding space employed by joint embedding approaches, does not sufficiently account for both sharable and private information. Such alignment risks transferring source-private $C'_s$, which has no relevant counterpart in the target $C_t$, thereby introducing errors or ``noise" into the target and leading to the inappropriate transfer of information. How to explicitly build effective transfer under such partial transferability remains a key challenge. It requires the alignment of domain space and identification of preferences for knowledge sharing.

In addition, after identifying the shared portion between the source and target tasks, there is a risk of overlooking crucial target-private information $C'_t$ and $D'_t$. Specifically, $C'_t$ and $D'_t$ represent labels and domain information that are unique to the target and do not appear in the source. Solely aligning domain knowledge $D_s$ and optimizing knowledge transfer in the source label space $C_s$ means that may ignore any private, target-specific knowledge. This leads to a deficiency in utilizing target-private information within the shared knowledge.

Therefore, the knowledge transfer method applied in this paper is expected not only to explicitly discover a domain-agnostic space and optimize knowledge transfer in the source label space but also to incorporate complementary information into the target space.

\subsection{Overall Architecture}

\begin{figure}[!t]
\centering
\includegraphics[width=\columnwidth]{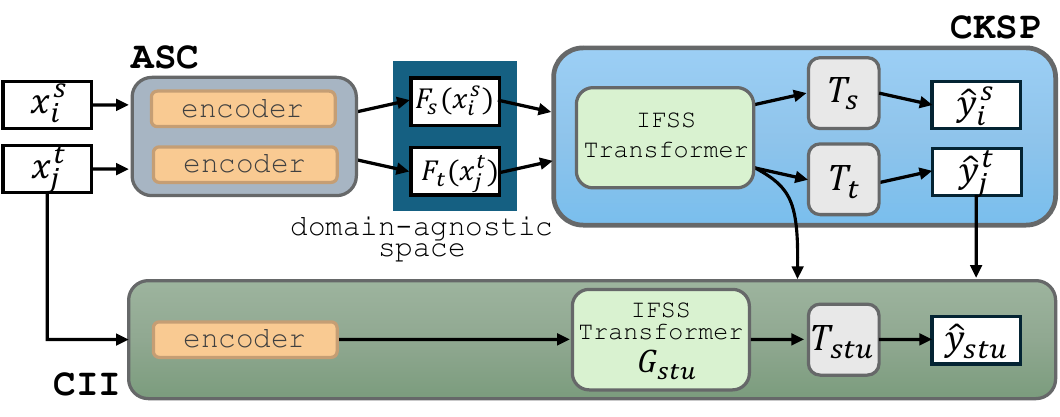}
\caption{
Our method comprises three components: 	Alignment of Spectral Characteristics (ASC), 	Cross-scene Knowledge Sharing Preference (CKSP), and Complementary Information Integration (CII). ASC addresses spectral discrepancies caused by different HSI sensors by encoding the input data $x^s_{i}$ and $x^t_{j}$ into a domain-agnostic space using an adversarial training scheme with two transformation functions, $F_s$ and $F_t$. CKSP resolves semantic mismatches due to category correspondence issues by identifying shared semantics through an introduced Source Similarity Mechanism, ensuring adaptive transfer. CKSP includes an Interaction Fusion Spatial-Spectral (IFSS) Transformer learner and task heads $T_{s}$ and $T_{t}$. Finally, CII tackles the deficiency in utilizing target-private information within shared knowledge, caused by domain and semantic mismatches, by introducing complementary extraction and distillation integration to build the final integrated target student model with $G_{stu}$ and $T_{stu}$.
}
\label{overall}
\end{figure}

Our goal is to transfer knowledge from the source to the target scene, thereby enhancing the target scene's discriminative power. As discussed above, we face three main challenges: spectral discrepancies stemming from different HSI sensors, mismatched semantics caused by the lack of category correspondence, and a deficiency in leveraging target-private information within shared knowledge. To address these challenges, our model comprises three components: the Alignment of Spectral Characteristics (ASC), Cross-scene Knowledge Sharing Preference (CKSP), and Complementary Information Integration (CII) modules, as shown in Fig. \ref{overall}.

ASC addresses spectral discrepancies by mapping the source and target distributions into a domain-agnostic space, ensuring consistency across different HSI sensors. CKSP handles semantic mismatches by employing a source similarity mechanism to identify semantic similarities between the source and target tasks. This mechanism determines the importance of semantic sharing preferences for each source sample. CII resolves the inadequate utilization of target-private information by leveraging complementary extraction and distillation integration modules, thereby ensuring comprehensive knowledge transfer and enhanced discrimination in the target scene.


 
\subsection{Alignment of Spectral Characteristics}
\begin{figure}[!t]
\includegraphics[width=\columnwidth]{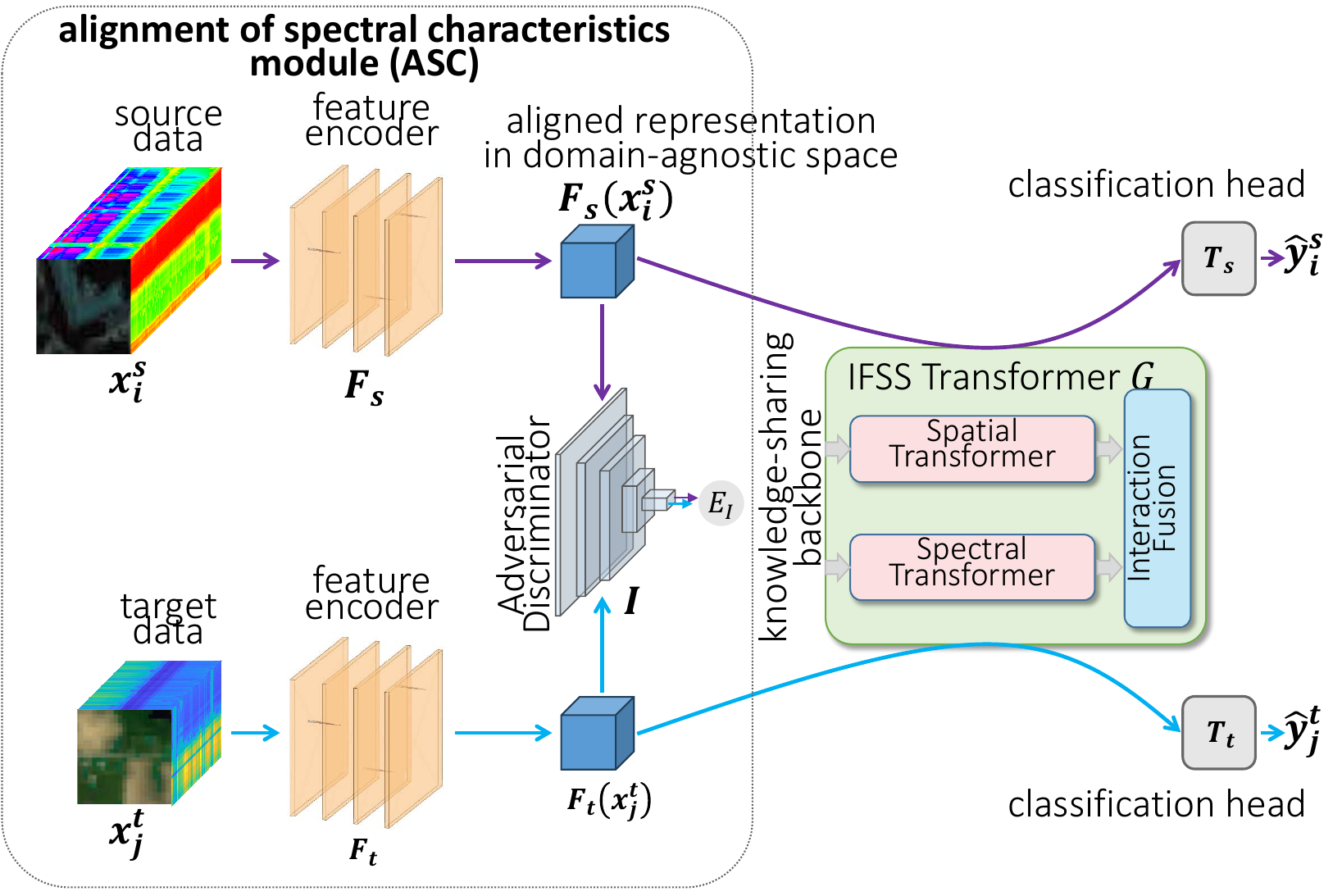}
\caption{ASC utilizes two transformation encoders, $F_s$ and $F_t$, to project the input data $x^s_{i}$ and $x^t_{j}$ into a domain-agnostic space via an adversarial discriminator $I$. The IFSS transformer network $G$ is shared by both the source and target data, while $T_s$ and $T_t$ serve as the classification heads for the source and target, respectively. }
\label{ASC}
\end{figure}

The different criteria for HSI data acquisition results in spectral discrepancies, which manifested as spectral shifts and variations  in the number of bands across different types of HSI sensors.
Alignment of Spectral Characteristics (ASC) employs two transformation functions $F_s$ and $F_t$ that project input data $x^s_{i}$ and $x^t_{j}$ from different domains into the domain-agnostic space $D$, as shown in Fig \ref{ASC}.

To align data distributions from different domains, adversarial learning serves as an effective strategy to bridge the domain gap, reducing feature-distribution discrepancies between the source and target domains. Therefore, we introduce adversarial learning to enforce the projected features into the domain-agnostic space $D$.
The loss term $E_{I}$ in Eq. (\ref{Intersection_spectral_task}) utilizes adversarial discriminator $I$, which is designed to match the spectral discrepancy adversarially:


\begin{equation}
\label{Intersection_spectral_task}
\begin{split}
E_{I} = -\mathbb{E}_{x^s_{i}\sim p}\log I(F_s(x^s_{i})) -\mathbb{E}_{x^t_{j}\sim q}\log(1- I(F_t(x^t_{j}))  
\end{split}
\end{equation}
where $p$ represents the source distribution while $q$ indicates the target distribution. 

To ensure effective knowledge transfer, the features from both the source and target domains should retain discriminative power. Hence, the representations in the domain-agnostic space are passed to a shared network learner, $G$, and then separately fed into two specific task heads, $T_s$ for the source and $T_t$ for the target HSI classification. The respective losses are denoted as $E_{T_s}$ and $E_{T_t}$. 

Given the effectiveness of the spatial-spectral Transformer in extracting both spatial and spectral information from hyperspectral images \cite{scheibenreif2023masked}, we adapt this model by employing an information-fusion scheme \cite{huo2024center} tailored to our task. As shown in Fig. \ref{ASC}, the inputs $x \in D_s$ and $x \in D_t$ are mapped into a domain-agnostic space $D$ via the adversarial discriminator $I$. Meanwhile, $G$ denotes the shared information-fusion spatial-spectral (IFSS) network that transfers information from the source to the target. Two MLP heads, $T_s$ and $T_t$, produce the respective classification outcomes.

The training of ASC can be described as a minimax game:
\begin{equation}
\label{minimax}
\begin{split}
   E_{\text{ASC}} = \max_I\min_{F_s,F_t,G,T_s,T_t}E_{T_s}+E_{T_t}-\lambda E_I\\
\end{split}
\end{equation}
where $\lambda$ serves as a hyperparameter to balance transferability term $E_I$ and discriminability terms $E_{T_s}, E_{T_t}$.
The gradient reversal layer, as proposed by Ganin et al. \cite{ganin2016domain}, inverts the gradient between $I$ and $T_s, T_t$, facilitating end-to-end optimization of all components.


\subsection{Cross-scene Knowledge Sharing Preference}
\begin{figure}[!t]
\includegraphics[width=\columnwidth]{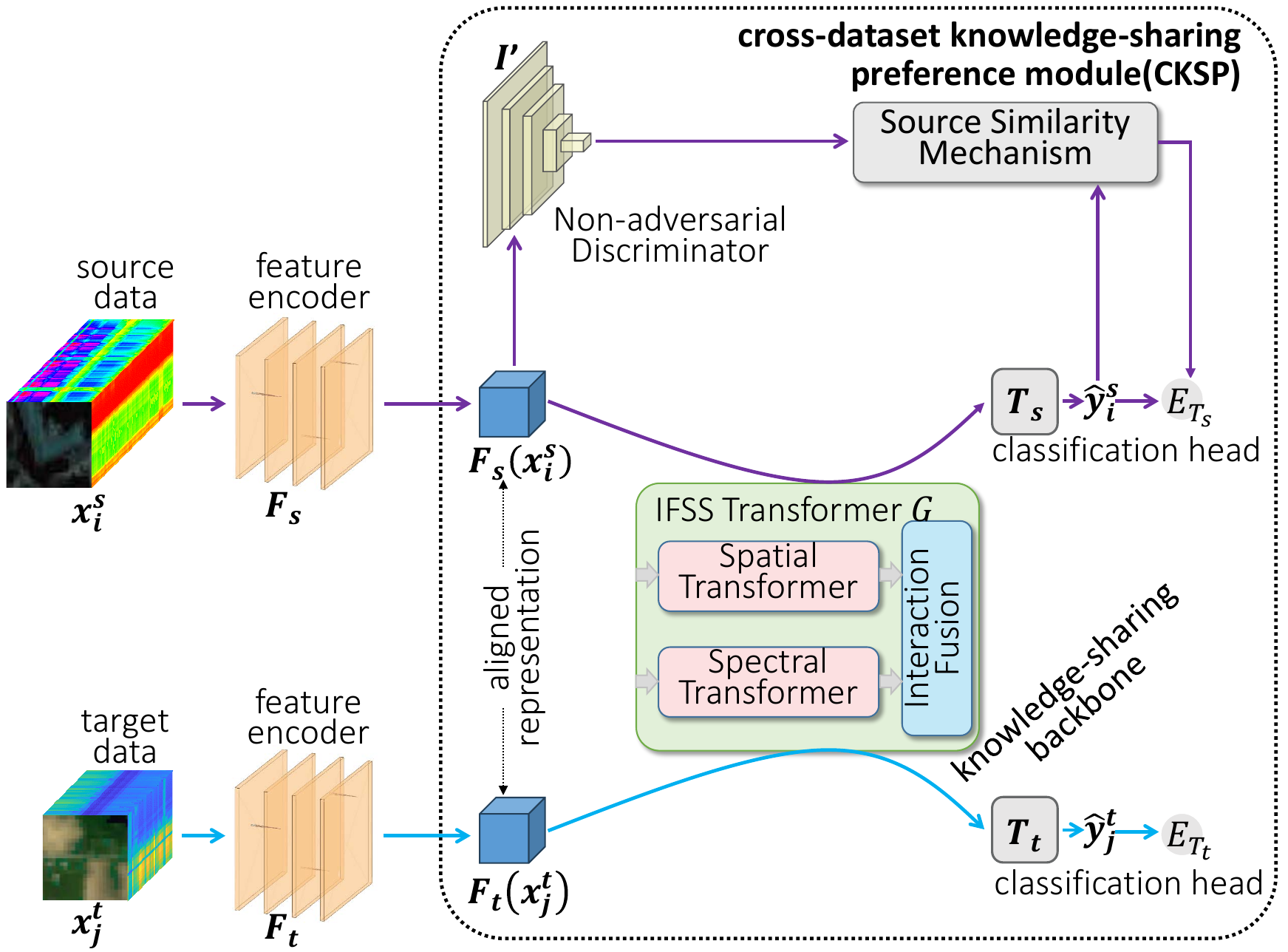}
\caption{CKSP leverages a Source Similarity Mechanism to determine the importance of each source sample during the knowledge transfer  from source to target. This mechanism gives greater weight $\omega^s(x)$ to source samples with higher semantic similarity though entropy uncertainty $H(\widehat{y^s})$ and domain similarity ${I'}(F_s(x^s))$ to enhance knowledge sharing.}
\label{CKSP}
\end{figure}

In general scenarios, categories from different tasks are semantically mismatched, making traditional domain adaptation methods ineffective. This semantic mismatch leads to a lack of one-to-one correspondence between the source and target domains. Therefore, it is crucial to determine which parts of the source semantic information will benefit the target task by identifying the shared portions between the source and target domains to enable effective knowledge transfer. To address this challenge, we introduce a Source Similarity Mechanism (SSM) that establishes sample-level semantic correspondence to facilitate knowledge transfer, as illustrated in Fig. \ref{CKSP}.

 The SSM employs semantic similarity to correlate the source with the target task. To identify which source information most benefits the target task, the SSM assigns an importance score to each source sample during knowledge transfer. The more important source samples show a higher semantic similarity to the target task, while less important samples exhibit a lower semantic similarity.

In the proposed SSM, we use $\omega^s(x^s_{i})$ to indicate the semantic similarity of a source sample 
to the target scene. To optimize knowledge transfer, $\omega^s(x^s_{i})$ should be higher for samples within the intersectional semantic space $C_t \cap C_s$ and lower for those in the external space $C_s \setminus C_t$. Therefore, $\omega^s(x^s_{i})$ should satisfy the following inequality:
\begin{equation}
\label{weight}
\begin{split}
    \mathbb{E}_{x^s_{i}\sim p_{(C_s \cap C_t)}}\omega^s(x^s_{i}) &> 
    \mathbb{E}_{x^s_{i}\sim p_{(C_s \setminus C_t)}}\omega^s(x^s_{i}) 
\end{split}
\end{equation}

A discriminator for the source and target scenes can potentially be used as the indicator to satisfy the above property.
Since source-private data $C_s \setminus C_t$ can be easier to distinguish than semantically shared samples, which can result in higher discriminative confidence. Considering existing discriminator functions $I$ related to adversarial training can lack sufficient discriminatory ability, we introduce a non-adversarial discriminator $I'$, which determines domain similarity and is trained with the following loss function:
\begin{equation}
\label{Intersection_spectral_task_1}
\begin{split}
 E_{I'} = &-\mathbb{E}_{x^s_{i}\sim p} \log I'(F_s(x^s_{i}))\\ 
&-\mathbb{E}_{x^t_{j}\sim q} \log(1- I'(F_t(x^t_{j})))   
\end{split}
\end{equation}
To satisfy the inequality for $\omega^s(x^s_{i})$, we use the output of $I'$ to measure domain similarity. The function $I'$ identifies samples from the source domain as 1 and samples from the target domain as 0. The value of the discriminative probability ranges between 0 and 1. A smaller $I'$ value for a source sample indicates greater similarity to the target. Thus, we ensure that:
\begin{equation}
\label{similarity}
\mathbb{E}_{x^s_{i} \sim p_{(C_s \cap C_t)}} I'(F_s(x^s_{i})) < \mathbb{E}_{x^s_{i}\sim p_{(C_s \setminus C_t)}} I'(F_s(x^s_{i})) 
\end{equation}

Moreover, the inequality for $\omega^s(x^s_{i})$ can be linked to entropy uncertainty to represent the semantic similarity. Source-private data are labeled uniquely in the semantic space $C_s$, whereas semantically shared samples in $C_t \cap C_s$ have different semantic labels. Thus, it can be assumed that source-private samples exhibit lower semantic uncertainty. Because entropy is an effective metric for characterizing uncertainty, we use entropy to assess the uncertainty of predictions for $\hat{y}$. The entropy of source samples from $C_s \setminus C_t$ is expected to be lower than that of source samples from $C_s \cap C_t$. Therefore, the source samples satisfy the following condition:
\begin{equation}
\label{entropy}
\mathbb{E}_{x^s_{i}\sim p_{(C_s \cap C_t)}}H(\widehat{y^s_{i}}) > \mathbb{E}_{x^s_{i}\sim p_{(C_s \setminus C_t)}}H(\widehat{y^s_{i}})
\end{equation}
where $\widehat{y^s_{i}}$ represents the predicted probability by the source classification head and $H$ denotes the entropy operator.

Based on the preceding analysis obtaining Eq. \eqref{similarity} and Eq. \eqref{entropy}, we can define the following $\omega^s(x^s_{i})$ to satisfy Eq. \eqref{weight}:

\begin{equation}
\label{weight_s}
\omega^s(x^s_{i}) = \frac{H(\widehat{y^s_{i}})}{\log\left|C_s\right|} - I'(F_s(x^s_{i}))
\end{equation}
where $\log|C_s|$ serves as the entropy normalization factor. This semantic similarity determines the significance of each source sample during knowledge sharing. By assessing semantic similarity, we can identify the most relevant source information, emphasizing the shared portion and thereby improving the effectiveness of knowledge sharing.

The value of $\omega^s(x^s_{i})$ represents the importance of each source sample when sharing knowledge from source to target. Specifically, samples with higher $\omega^s(x^s_{i})$ values are considered more important due to their greater semantic similarity with the target domain. Consequently, the classification loss for the source task $E_{T_s}$ in Eq. \eqref{minimax} should be updated to reflect this importance:
\begin{equation}
\label{error_Source_class_2}
E_{T_s} = \mathbb{E}_{(x^s_{i}, y^s_{i})\sim p}\omega^s(x^s_{i})L(y^s_{i}, T_s(G(F_s(x^s_{i}))))
\end{equation}
where $L$ denotes sample-wise cross-entropy loss. Comparatively, the target classification loss is defined as:
\begin{equation}
\label{error_Target_class_1}
E_{T_t} = \mathbb{E}_{(x^t_{j}, y^t_{j})\sim q}L(y^t_{j}, T_t(G(F_t(x^t_{j}))))
\end{equation}
Therefore, the CKSP  with ASC loss can be described as:
\begin{equation}
\label{SSM_loss}
E_{\text{ASC-}\text{CKSP}} = E_{I'} + E_{\text{ASC}}
\end{equation}
where $E_{T_s}$ and $E_{T_t}$ in $E_\text{ASC}$ are specified as in Eq. \eqref{error_Source_class_2}, Eq. \eqref{error_Target_class_1} and Eq. \eqref{minimax}, and $E_{I'}$ is defined as in Eq. \eqref{Intersection_spectral_task_1}.

\subsection{Complementary Information Integration}

\begin{figure}[!t]
\includegraphics[width=\columnwidth]{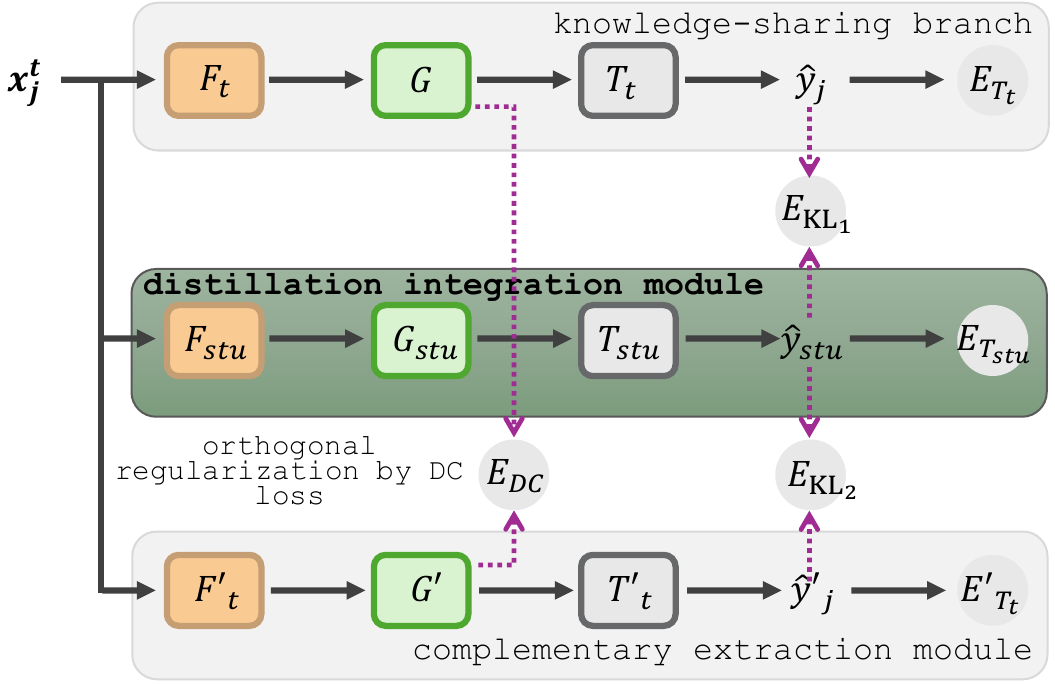}
\caption{
CII contains Complementary Extraction (CE) and Distillation
Integration (DI)  modules to ensure the complete information of the target data.
We employ Partial Distance Correlation $E_{DC}$ to find the target-private information to identify target-private information that supplements the shared information between the source and target. 
To further guarantee the completeness of the target information, we introduce a distillation integration method, with $E_{KL1}$ learning from shared information and $E_{KL2}$ learning from complementary information, thereby enhancing the representation of the entire target scene.
}
\label{CII}
\end{figure}

Due to spectral discrepancies and semantic mismatches, aligning the spectral domain and knowledge-sharing preferences allows the model to focus solely on the shared portions of target information. However, this overlooks or diminishes target-private information, including unique domain characteristics and task-specific category details. To ensure the extraction of discriminative features 
 for complete set of target information, we introduce Complementary Information Integration (CII), consisting of Complementary Extraction (CE) and Distillation Integration (DI) modules, as shown in Fig. \ref{CII}.

We propose the CE module to represent the target-private information. This target-private information supplements the shared information between source and target, and each type of information remains independent. The Partial Distance Correlation \cite{zhen2022versatile} is applied as an objective to minimize the association between these two information:
\begin{equation}
\label{dc_target}
E_{\text{DC}} = \mathbb{E}_{x\sim q}dCor(G(F_t(x)), G'(F'_t(x)))
\end{equation}
where $F_t', G'$ and $T_t'$ denote the separate components capturing target-private (complementary) information $G'(F_t'(x)))$, which is enforced to be orthogonal to the shared information $G(F_t(x)))$.

To ensure the completeness of the target information, we integrate the shared and target-private information. Therefore, we propose a DI module to enhance the representation of the entire target scene. In our approach, we employ two teacher models and one student model to merge the complementary and shared information of the target data. Specifically, the DI module has two teachers: one extracts shared information $G(F_t(x)))$ and the other extracts target-private information $G'(F_t'(x)))$. The student model, consisting of $G_{stu}$ and $T_{stu}$, then integrates the information from both teachers. Additionally, we employ reverse distillation \cite{li2022shadow} to iteratively update the student-aware teacher model, thereby reducing the discrepancies between the teacher and student models, which enhances the efficiency of the distillation process.

To train the DI module, the KD loss for the shared teacher and the student is represented by $E_{\text{KL}_1}$ in Eq. \eqref{kt_1}. Similarly, $E_{\text{KL}_2}$ in Eq. \eqref{kt_2} represents the KD loss for the target-private teacher and the student. By learning from both the shared teacher and the target-private teacher, the student model effectively integrates these two types of information, thereby ensuring the completeness of the target information.
\begin{equation}
\label{kt_1}
\begin{split}
E_{\text{KL}_1} & = \mathbb{E}_{x\sim q} \big [D_{\text{KL}}(T_{stu}(G_{stu}(F_t(x))), T_t(G(F_t(x))))\\ 
&+ D_{\text{KL}}( T_t(G(F_t(x))), T_{stu}(G_{stu}(F_t(x)))) \big ]
\end{split}
\end{equation}
\begin{align}
E_{\text{KL}_2} & = \mathbb{E}_{x\sim q} \big [ D_{\text{KL}}(T_{stu}(G_{stu}(F_t(x))), T'_t(G'(F_t'(x)))) \nonumber \\ 
&+ D_{\text{KL}}( T'_t(G'(F_t'(x))), T_{stu}(G_{stu}(F_t(x))))  \big ] \label{kt_2}
\end{align}
where $D_{\text{KL}}$ denotes KL divergence. The DI process, which integrates the shared information between source and target with the target-private information into a student model, can be outlined as follows:

\begin{equation}
\label{kl}
E_{\text{KL}} = E_{\text{KL}_1} +E_{\text{KL}_2} 
\end{equation}
The student model is trained to capture complete information across the entire target space.

To ensure that features extracted by the CE and the DI modules retain discriminative power, we apply classification losses $E'_{T_t}$ for CE module and  $E_{T_{stu}}$ for DI module as follows:
\begin{equation}
\label{error_Target_class}
E'_{T_t} = \mathbb{E}_{(x, y)\sim q}L(y, T'_t(G'(F_t'(x))))
\end{equation}
\begin{equation}
\label{error_Target_class}
E_{T_{stu}} = \mathbb{E}_{(x, y)\sim q}L(y, T_{stu}(G_{stu}(F_t(x))))
\end{equation}
As illustrated in Fig. \ref{CII}, the training loss for the CII module is expressed as:
\begin{equation}
\label{error_Target_all}
E_{\text{CII}} = \alpha E_{\text{KL}} + \beta E'_{T_t} + \gamma E_{T_{stu}} + E_{\text{DC}}
\end{equation}
Finally, the overall loss of our model can be written as:
\begin{equation}
\label{class_all}
E = E_{\text{ASC-CKSP}} + E_{\text{CII}}
\end{equation}

\section{Experiments}
We evaluate the CKI framework on scenarios in which the source and target spaces exhibit different spectral characteristics and distinct label spaces across several datasets. Next, we investigate the impact of various components and hyperparameter sensitivity through an ablation study.
\subsection{Experimental Setup}
In this subsection, we describe the implementation details, the datasets used, and the evaluation criteria.

\subsubsection{Datasets}

\begin{table}[]
\caption{The detailed categories information for Indian Pines, Pavia University and Houston 2013 datasets.}
\begin{tabular}{c|c|c}
\hline
\bottomrule
 Indian Pines & Pavia University & Houston 2013 \\ \hline
 Alfalfa & Asphalt & Health grass  \\
  Corn-notill & Meadows & Stressed grass  \\
 Corn-mintill & Gravel & Synthetic grass\\
  Corn & Trees & Tress \\
 Grass-pasture & Painted metal sheets  & Soil \\
 Grass-trees & Bare Soil & Water \\
 Grass-pasture-mowed & Bitumen & Residential \\
 Hay-windrowed & Self-Blocking Bricks & Commercial \\
 Oats & Shadows & Road \\
 Soybean-notill &  & Highway \\
 Soybean-mintill &  & Railway \\
 Soybean-clean &  & Parking lot 1 \\
 Wheat &  & Parking lot 2 \\
 Woods &  & Tennis court\\
 Buildings-Grass-Trees-Drives &  & Running track  \\
 Stone-Steel-Towers &  &  \\ \hline
  \bottomrule
\end{tabular}
\end{table}
As illustrated in Table I, the Indian Pines, Pavia University, and Houston 2013 datasets were acquired using different sensors, each with a unique wavelength range and number of bands. These three datasets contain distinct categories, as illustrated in Table II. Furthermore, any of these three datasets can serve as either source or target data. Note that most categories do not exhibit a one-to-one correspondence between the source and target. Additionally, we follow prior work as a baseline \cite{mohamed2023factoformer,hong2021spectralformer,ibanez2022masked} for our full training and testing sets. In our default setting, the testing set remains the same. To evaluate knowledge transfer performance under limited target samples, ten training samples for each category are randomly selected from the full training set in the target dataset.

The Indian Pines dataset, captured by the AVIRIS sensor over northwestern Indiana, comprises $145\times145$ pixels with 224 spectral reflectance bands within the $0.4–2.5 \mu$m wavelength range.
The provided ground truth is categorized into sixteen classes, though not all are mutually exclusive. Additionally, to mitigate water absorption effects, the number of bands has been reduced to 200 \cite{wang2017locality,nalepa2019validating, li2016hyperspectral}. 



The Pavia University scenes were captured by the ROSIS sensor during a flight campaign over Pavia University in northern Italy. This dataset contains 103 spectral bands and encompasses an area of $610\times610$ pixels with a geometric resolution of 1.3 meters. The image ground truths distinguish nine unique classes.

The Houston 2013 dataset, collected over the University of Houston campus and the surrounding urban area for the 2013 IEEE GRSS Data Fusion Contest, spans $349\times1905$ pixels at a spatial resolution of 2.5 meters. This hyperspectral imaging signal comprises 144 bands ranging from 0.38 to 1.05 $\mu$m, and is categorized into 15 distinct classes.
\subsubsection{Evaluation Details}
We use Overall Accuracy (OA), Average Accuracy (AA), and the Kappa coefficient ($\kappa$) for evaluation. OA calculates the ratio of correctly predicted observations to the total observations, offering a straightforward measure of the model's overall performance. AA computes the mean accuracy per class, ensuring that the model's performance is evenly assessed across all classes. $\kappa$ measures the agreement between model predictions and actual classification results \cite{li2024scformer}.
\subsubsection{Implementation details}

The CKI framework was implemented using PyTorch 1.12.0 in Python 3.8.16 with Intel(R) Xeon(R) E-2288G with 64-GB RAM with an NVIDIA Quadro RTX 6000 24 GB graphics processing unit (GPU). All the source training data will be employed while only 10 samples for each category in target training data will be used to train the model. In addition, we used a batch size of 64 with the Adam optimizer. The learning rate was set to 5e-4 with a weight decay 5e-3, with a $\gamma$ factor of 0.9 every 1/10 of the epochs was used. We used warm-up strategy to avoid local minima and stabilize the training, where $\lambda = min(1, \text{current epoch}/\text{warmup epochs})$. Here, we set the warmup epochs to 20. Hyperparameters $\alpha$, $\beta$ and $\gamma$ are all set to 1 in Eq. \eqref{error_Target_all}. In addition, we employed convolutional modules were used to build the transformation functions $F_s$, $F_t$, $F_{t’}$ and Domain Discriminators $I$, $I'$. Additionally, we used MLP modules for $T_s$, $T_t$ and $T_{stu}$ as classification heads. To enhance the exploration of spectral and spatial information, we employ both the spectral transformer and the spatial transformer, similar to the Masked SST \cite{scheibenreif2023masked}. We then fuse these two types of information via a 2×2 simplified fusion matrix \cite{huo2024center} without introducing the center patch of HSI as an extra input for the components $S$, $S'$ and $S_{stu}$. The specific architecture of IFSS transformer is provided in the Appendix section. 

\subsection{Knowledge Transfer Results}  
\textbf{Compared Methods:} 
Our proposed methodology, CKI, was compared with an array of established knowledge transfer methodologies. Among these were a Multi-Task Learning (MTL) \cite{lee2018cross}, domain alignment through the Universal Adaptation Network (UAN) \cite{you2019universal}, and KD-based ensemble learning mechanisms, including Feature Fusion Learning (FFL) \cite{kim2021feature} and On-the-fly Native Ensemble (ONE) \cite{zhu2018knowledge}. Additionally, strategies focused on parameter-efficient knowledge transfer--- Finetune \cite{lee2022exploring} and Adapter \cite{houlsby2019parameter}—-were also evaluated. To ensure a fair comparison, all methods, including ours, employed the same backbone architecture.

\textbf{Classification Results:} The classification results are presented in Tables \ref{tab:comparison-indianpian-to-huston}, \ref{tab:comparison-huston-to-pavia}, and \ref{tab:comparison-huston-to-indianpian}, showing the performance of our method compared to previous knowledge transfer methods. The baseline represents the supervised learning outcomes of 10 training samples per category in the target dataset.

\begin{table*}[]
\centering
\caption{The OA, AA, $\kappa$ values (\%) of knowledge transfer from Pavia University to Houston and from IndianPian to Houston}
\label{tab:comparison-indianpian-to-huston}
\begin{adjustbox}{width=\linewidth}
\setlength{\tabcolsep}{0.5mm}{
\begin{tabular}{c|c|ccccccc|ccccccc}
\hline
\bottomrule
\multirow{2}{*}{Class} & \multirow{2}{*}{Baseline} & \multicolumn{7}{c|}{Pavia University$\rightarrow$Houston} & \multicolumn{7}{c}{Indian Pines$\rightarrow$Houston} \\ \cline{3-16} 
&                           & \multicolumn{1}{c|}{MTL} & \multicolumn{1}{c|}{UAN} & \multicolumn{1}{c|}{ONE} & \multicolumn{1}{c|}{FFL} & \multicolumn{1}{c|}{Adaptor} & \multicolumn{1}{c|}{Finetune} & Ours & \multicolumn{1}{c|}{MTL} & \multicolumn{1}{c|}{UAN} & \multicolumn{1}{c|}{ONE} & \multicolumn{1}{c|}{FFL} & \multicolumn{1}{c|}{Adaptor} & \multicolumn{1}{c|}{Finetune} & Ours \\ \hline
   Healthy grass& 74.55 & \multicolumn{1}{c|}{70.28} & \multicolumn{1}{c|}{68.85} & \multicolumn{1}{c|}{63.53} & \multicolumn{1}{c|}{70.66} & \multicolumn{1}{c|}{74.55} & \multicolumn{1}{c|}{\color{blue}{76.26}} & \color{red}{78.63} & \multicolumn{1}{c|}{75.59} & \multicolumn{1}{c|}{71.89} & \multicolumn{1}{c|}{\color{blue}{76.92}} & \multicolumn{1}{c|}{76.35} & \multicolumn{1}{c|}{69.99} & \multicolumn{1}{c|}{73.03} & \color{red}{77.40}  \\ \hline
                 Stressed grass& 89.85 & \multicolumn{1}{c|}{92.67} & \multicolumn{1}{c|}{95.68} & \multicolumn{1}{c|}{\color{red}{97.37}} & \multicolumn{1}{c|}{\color{blue}{97.27}} & \multicolumn{1}{c|}{94.83} & \multicolumn{1}{c|}{92.95} & 9643 & \multicolumn{1}{c|}{95.86} & \multicolumn{1}{c|}{95.21} & \multicolumn{1}{c|}{\color{blue}{96.90}} & \multicolumn{1}{c|}{\color{red}{97.65}} & \multicolumn{1}{c|}{95.39} & \multicolumn{1}{c|}{92.01} & 94.55  \\ \hline
                 Synthetic grass& 54.65 & \multicolumn{1}{c|}{70.89} & \multicolumn{1}{c|}{\color{blue}{75.25}} & \multicolumn{1}{c|}{74.65} & \multicolumn{1}{c|}{68.32} & \multicolumn{1}{c|}{73.27} & \multicolumn{1}{c|}{48.71} & \color{red}{78.81} & \multicolumn{1}{c|}{60.20} & \multicolumn{1}{c|}{64.75} & \multicolumn{1}{c|}{74.06} & \multicolumn{1}{c|}{\color{blue}{79.80}} & \multicolumn{1}{c|}{75.64} & \multicolumn{1}{c|}{\color{red}{80.79}} & 75.25 \\ \hline
                 Trees&91.66  & \multicolumn{1}{c|}{\color{blue}{97.06}} & \multicolumn{1}{c|}{91.57} & \multicolumn{1}{c|}{96.31} & \multicolumn{1}{c|}{96.88} & \multicolumn{1}{c|}{93.66} & \multicolumn{1}{c|}{96.02} & \color{red}{98.11} & \multicolumn{1}{c|}{89.20} & \multicolumn{1}{c|}{96.12} & \multicolumn{1}{c|}{\color{red}{97.25}} & \multicolumn{1}{c|}{\color{blue}{96.59}} & \multicolumn{1}{c|}{95.74} & \multicolumn{1}{c|}{8873} &94.22  \\ \hline
                Soil& 99.53  & \multicolumn{1}{c|}{99.34} & \multicolumn{1}{c|}{97.82} & \multicolumn{1}{c|}{99.43} & \multicolumn{1}{c|}{96.31} & \multicolumn{1}{c|}{98.20} & \multicolumn{1}{c|}{\color{red}{99.62}} & \color{red}{99.62} & \multicolumn{1}{c|}{99.43} & \multicolumn{1}{c|}{\color{red}{100.00 }} & \multicolumn{1}{c|}{98.86} & \multicolumn{1}{c|}{99.15} & \multicolumn{1}{c|}{98.11} & \multicolumn{1}{c|}{\color{blue}{99.81}} & 99.53  \\ \hline
                 Water & 77.62 & \multicolumn{1}{c|}{74.13} & \multicolumn{1}{c|}{81.82} & \multicolumn{1}{c|}{81.82} & \multicolumn{1}{c|}{\color{blue}{82.52}} & \multicolumn{1}{c|}{69.23} & \multicolumn{1}{c|}{70.63} & \color{red}{84.62} & \multicolumn{1}{c|}{\color{blue}{83.22}} & \multicolumn{1}{c|}{77.62} & \multicolumn{1}{c|}{81.12} & \multicolumn{1}{c|}{80.42} & \multicolumn{1}{c|}{69.23} & \multicolumn{1}{c|}{65.03} & \color{red}{86.01} \\ \hline
                Residential& 78.82  & \multicolumn{1}{c|}{\color{red}{88.25}} & \multicolumn{1}{c|}{\color{blue}{86.75}} & \multicolumn{1}{c|}{79.48} & \multicolumn{1}{c|}{86.01} & \multicolumn{1}{c|}{84.24} & \multicolumn{1}{c|}{86.38} & 85.35  & \multicolumn{1}{c|}{\color{red}{84.24}} & \multicolumn{1}{c|}{83.77} & \multicolumn{1}{c|}{83.02} & \multicolumn{1}{c|}{\color{blue}{83.86}} & \multicolumn{1}{c|}{77.71} & \multicolumn{1}{c|}{72.20} & 80.88  \\ \hline
                Commercial & \color{red}{83.38}   & \multicolumn{1}{c|}{68.76} & \multicolumn{1}{c|}{68.28} & \multicolumn{1}{c|}{72.27} & \multicolumn{1}{c|}{73.12} & \multicolumn{1}{c|}{68.38} & \multicolumn{1}{c|}{76.07} & \color{blue}{77.87} & \multicolumn{1}{c|}{77.59} & \multicolumn{1}{c|}{78.44} & \multicolumn{1}{c|}{76.07} & \multicolumn{1}{c|}{68.57} & \multicolumn{1}{c|}{\color{blue}{85.28}} & \multicolumn{1}{c|}{\color{red}{87.27}} & 82.81\\ \hline   
                Road & \color{blue}{82.34} & \multicolumn{1}{c|}{71.01} & \multicolumn{1}{c|}{69.78} & \multicolumn{1}{c|}{\color{red}{84.23}} & \multicolumn{1}{c|}{62.04} & \multicolumn{1}{c|}{\color{blue}{82.91}} & \multicolumn{1}{c|}{66.57} & 75.35 & \multicolumn{1}{c|}{63.36} & \multicolumn{1}{c|}{64.78} & \multicolumn{1}{c|}{79.23} & \multicolumn{1}{c|}{67.61} & \multicolumn{1}{c|}{\color{red}{86.87}} & \multicolumn{1}{c|}{78.94} & 58.64 \\ \hline       
                 Highway& 62.36 & \multicolumn{1}{c|}{\color{blue}{72.49}} & \multicolumn{1}{c|}{55.21} & \multicolumn{1}{c|}{56.56} & \multicolumn{1}{c|}{68.82} & \multicolumn{1}{c|}{48.84} & \multicolumn{1}{c|}{53.47} & \color{red}{85.04} & \multicolumn{1}{c|}{51.74} & \multicolumn{1}{c|}{\color{blue}{66.51}} & \multicolumn{1}{c|}{51.74} & \multicolumn{1}{c|}{66.12} & \multicolumn{1}{c|}{62.93} & \multicolumn{1}{c|}{55.50} & \color{red}{92.76}\\ \hline
                Railway& 70.97  & \multicolumn{1}{c|}{77.80} & \multicolumn{1}{c|}{80.74} & \multicolumn{1}{c|}{82.07} & \multicolumn{1}{c|}{82.45} & \multicolumn{1}{c|}{80.36} & \multicolumn{1}{c|}{\color{red}{86.81}} & \color{blue}{83.49} & \multicolumn{1}{c|}{73.72} & \multicolumn{1}{c|}{\color{blue}{80.83}} & \multicolumn{1}{c|}{\color{red}{86.15}} & \multicolumn{1}{c|}{78.08} & \multicolumn{1}{c|}{79.32} & \multicolumn{1}{c|}{78.56} & 79.70  \\ \hline
                 Parking Lot 1& 71.47 & \multicolumn{1}{c|}{71.18} & \multicolumn{1}{c|}{\color{blue}{72.43}} & \multicolumn{1}{c|}{64.07} & \multicolumn{1}{c|}{62.34} & \multicolumn{1}{c|}{70.32} & \multicolumn{1}{c|}{57.44} & \color{red}{75.60} & \multicolumn{1}{c|}{\color{blue}{73.20}} & \multicolumn{1}{c|}{68.20} & \multicolumn{1}{c|}{68.78} & \multicolumn{1}{c|}{66.57} & \multicolumn{1}{c|}{67.72} & \multicolumn{1}{c|}{58.50} & \color{red}{76.27}  \\ \hline
                 Parking Lot 2 & \color{red}{90.18} & \multicolumn{1}{c|}{87.37} & \multicolumn{1}{c|}{88.77} & \multicolumn{1}{c|}{86.67} & \multicolumn{1}{c|}{82.81} & \multicolumn{1}{c|}{78.95} & \multicolumn{1}{c|}{81.75} & \color{blue}{89.82} & \multicolumn{1}{c|}{84.21} & \multicolumn{1}{c|}{84.21} & \multicolumn{1}{c|}{85.96} & \multicolumn{1}{c|}{\color{blue}{89.47}} & \multicolumn{1}{c|}{72.63} & \multicolumn{1}{c|}{84.56} & 88.42\\ \hline
               Tennis Court  & 81.38   & \multicolumn{1}{c|}{81.78} & \multicolumn{1}{c|}{\color{blue}{90.28}} & \multicolumn{1}{c|}{78.54} & \multicolumn{1}{c|}{\color{red}{92.71}} & \multicolumn{1}{c|}{83.40} & \multicolumn{1}{c|}{76.11} & 84.21 & \multicolumn{1}{c|}{\color{blue}{90.28}} & \multicolumn{1}{c|}{88.07} & \multicolumn{1}{c|}{\color{red}{91.09}} & \multicolumn{1}{c|}{87.04} & \multicolumn{1}{c|}{73.68} & \multicolumn{1}{c|}{88.26} & 83.00\\ \hline
               Running Track  & 69.13  & \multicolumn{1}{c|}{64.27} & \multicolumn{1}{c|}{67.02} & \multicolumn{1}{c|}{75.48} & \multicolumn{1}{c|}{\color{blue}{78.01}} & \multicolumn{1}{c|}{57.29} & \multicolumn{1}{c|}{\color{red}{84.14}} & 67.02 & \multicolumn{1}{c|}{65.75} & \multicolumn{1}{c|}{71.88} & \multicolumn{1}{c|}{\color{red}{86.05}} & \multicolumn{1}{c|}{\color{blue}{79.70}} & \multicolumn{1}{c|}{65.33} & \multicolumn{1}{c|}{54.55} & 67.44 \\ \hline
               OA ($\uparrow$)  & 79.24  & \multicolumn{1}{c|}{\color{blue}{79.96}} & \multicolumn{1}{c|}{78.69} & \multicolumn{1}{c|}{79.42} & \multicolumn{1}{c|}{79.49} & \multicolumn{1}{c|}{78.52} & \multicolumn{1}{c|}{78.08} & \color{red}{84.64} & \multicolumn{1}{c|}{77.65} & \multicolumn{1}{c|}{79.85} & \multicolumn{1}{c|}{\color{blue}{81.73}} & \multicolumn{1}{c|}{80.45} & \multicolumn{1}{c|}{80.53} & \multicolumn{1}{c|}{77.87} & \color{red}{82.82} \\ \hline
                AA ($\uparrow$) & 78.53  & \multicolumn{1}{c|}{79.15} & \multicolumn{1}{c|}{79.35} & \multicolumn{1}{c|}{79.50} & \multicolumn{1}{c|}{\color{blue}{80.02}} & \multicolumn{1}{c|}{77.23} & \multicolumn{1}{c|}{76.86} & \color{red}{84.00} & \multicolumn{1}{c|}{77.84} & \multicolumn{1}{c|}{79.49} & \multicolumn{1}{c|}{\color{blue}{82.21}} & \multicolumn{1}{c|}{81.13} & \multicolumn{1}{c|}{78.37} & \multicolumn{1}{c|}{77.18} & \color{red}{82.46}\\ \hline
               $\kappa$ ($\uparrow$) & 77.52  & \multicolumn{1}{c|}{\color{blue}{78.26}} & \multicolumn{1}{c|}{76.88} & \multicolumn{1}{c|}{77.67} & \multicolumn{1}{c|}{77.76} & \multicolumn{1}{c|}{76.69} & \multicolumn{1}{c|}{76.21} & \color{red}{83.33} & \multicolumn{1}{c|}{75.77} & \multicolumn{1}{c|}{78.15} & \multicolumn{1}{c|}{\color{blue}{80.18}} & \multicolumn{1}{c|}{78.80} & \multicolumn{1}{c|}{78.84} & \multicolumn{1}{c|}{75.98} &\color{red}{81.37}   \\ \hline
\bottomrule
\end{tabular}}
\end{adjustbox}
\end{table*}

\begin{table*}[]
\centering
\caption{The OA, AA, $\kappa$ values (\%) of knowledge transfer from IndianPian to Pavia University and from Houston to Pavia University}
\label{tab:comparison-huston-to-pavia}
\begin{adjustbox}{width=\linewidth}
\setlength{\tabcolsep}{0.5mm}{
\begin{tabular}{c|c|ccccccc|ccccccc}
\hline
\bottomrule
\multirow{2}{*}{Class} & \multirow{2}{*}{Baseline} & \multicolumn{7}{c|}{Indian Pines$\rightarrow$Pavia University} & \multicolumn{7}{c}{Houston$\rightarrow$Pavia University} \\ \cline{3-16} 
&                           & \multicolumn{1}{c|}{MTL} & \multicolumn{1}{c|}{UAN} & \multicolumn{1}{c|}{ONE} & \multicolumn{1}{c|}{FFL} & \multicolumn{1}{c|}{Adaptor} & \multicolumn{1}{c|}{Finetune} & Ours & \multicolumn{1}{c|}{MTL} & \multicolumn{1}{c|}{UAN} & \multicolumn{1}{c|}{ONE} & \multicolumn{1}{c|}{FFL} & \multicolumn{1}{c|}{Adaptor} & \multicolumn{1}{c|}{Finetune} & Ours \\ \hline
                 Asphalt& 83.68 & \multicolumn{1}{c|}{78.66} & \multicolumn{1}{c|}{78.11} & \multicolumn{1}{c|}{79.70} & \multicolumn{1}{c|}{73.65} & \multicolumn{1}{c|}{\color{red}{85.42}} & \multicolumn{1}{c|}{79.74} & \color{blue}{84.80}  & \multicolumn{1}{c|}{77.62} & \multicolumn{1}{c|}{78.78} & \multicolumn{1}{c|}{\color{red}{87.06}} & \multicolumn{1}{c|}{85.03} & \multicolumn{1}{c|}{79.17} & \multicolumn{1}{c|}{\color{blue}{85.25}} & 80.82 \\ \hline
                 Meadows& 69.02 & \multicolumn{1}{c|}{69.50} & \multicolumn{1}{c|}{82.34} & \multicolumn{1}{c|}{63.72} & \multicolumn{1}{c|}{73.39} & \multicolumn{1}{c|}{81.57} & \multicolumn{1}{c|}{\color{red}{86.10}} & \color{blue}{84.51} & \multicolumn{1}{c|}{59.80} & \multicolumn{1}{c|}{74.50} & \multicolumn{1}{c|}{71.46} & \multicolumn{1}{c|}{68.47} & \multicolumn{1}{c|}{76.64} & \multicolumn{1}{c|}{\color{blue}{80.31}} & \color{red}{92.51}  \\ \hline
                 Gravel& 57.25 & \multicolumn{1}{c|}{73.72} & \multicolumn{1}{c|}{66.89} & \multicolumn{1}{c|}{73.28} & \multicolumn{1}{c|}{71.46} & \multicolumn{1}{c|}{\color{red}{87.99}} & \multicolumn{1}{c|}{68.32} & \color{blue}{77.63} & \multicolumn{1}{c|}{\color{blue}{74.93}} & \multicolumn{1}{c|}{73.22} & \multicolumn{1}{c|}{71.35} & \multicolumn{1}{c|}{67.27} & \multicolumn{1}{c|}{73.39} & \multicolumn{1}{c|}{63.47} & \color{red}{84.90}  \\ \hline
                Trees & \color{blue}{90.01} & \multicolumn{1}{c|}{81.73} & \multicolumn{1}{c|}{82.49} & \multicolumn{1}{c|}{84.51} & \multicolumn{1}{c|}{\color{red}{91.14}} & \multicolumn{1}{c|}{86.44} & \multicolumn{1}{c|}{87.81} & 84.99 & \multicolumn{1}{c|}{87.23} & \multicolumn{1}{c|}{86.47} & \multicolumn{1}{c|}{\color{blue}{90.18}} & \multicolumn{1}{c|}{88.32} & \multicolumn{1}{c|}{\color{red}{94.71}} & \multicolumn{1}{c|}{82.66} & 87.12  \\ \hline
                Painted metal sheets & \color{blue}{99.01}  & \multicolumn{1}{c|}{94.34} & \multicolumn{1}{c|}{96.32} & \multicolumn{1}{c|}{87.78} & \multicolumn{1}{c|}{97.21} & \multicolumn{1}{c|}{96.77} & \multicolumn{1}{c|}{\color{red}{99.10}} & 97.66 & \multicolumn{1}{c|}{96.86} & \multicolumn{1}{c|}{95.87} & \multicolumn{1}{c|}{98.38} & \multicolumn{1}{c|}{97.12} & \multicolumn{1}{c|}{\color{blue}{98.65}} & \multicolumn{1}{c|}{90.57} & 97.04 \\ \hline
                 Bare Soil& 55.51 & \multicolumn{1}{c|}{79.31} & \multicolumn{1}{c|}{69.79} & \multicolumn{1}{c|}{\color{blue}{85.74}} & \multicolumn{1}{c|}{50.79} & \multicolumn{1}{c|}{79.18} & \multicolumn{1}{c|}{65.46} & \color{red}{93.09}& \multicolumn{1}{c|}{\color{red}{91.84}} & \multicolumn{1}{c|}{55.03} & \multicolumn{1}{c|}{54.64} & \multicolumn{1}{c|}{67.98} & \multicolumn{1}{c|}{41.23} & \multicolumn{1}{c|}{61.70} & \color{blue}{74.04}  \\ \hline
                 Bitumen& 68.50 & \multicolumn{1}{c|}{70.03} & \multicolumn{1}{c|}{69.42} & \multicolumn{1}{c|}{58.00} & \multicolumn{1}{c|}{\color{red}{72.17}} & \multicolumn{1}{c|}{56.57} & \multicolumn{1}{c|}{68.60} & \color{blue}{71.36} & \multicolumn{1}{c|}{\color{blue}{75.74}} & \multicolumn{1}{c|}{74.21} & \multicolumn{1}{c|}{67.99} & \multicolumn{1}{c|}{60.35} & \multicolumn{1}{c|}{69.22} & \multicolumn{1}{c|}{64.93} & \color{red}{79.31}  \\ \hline
                Self-Blocking Bricks& \color{red}{95.04}  & \multicolumn{1}{c|}{93.61} & \multicolumn{1}{c|}{\color{red}{95.30}} & \multicolumn{1}{c|}{89.80} & \multicolumn{1}{c|}{\color{red}{95.30}} & \multicolumn{1}{c|}{86.95} & \multicolumn{1}{c|}{93.01} & 93.67 & \multicolumn{1}{c|}{91.08} & \multicolumn{1}{c|}{92.24} & \multicolumn{1}{c|}{93.64} & \multicolumn{1}{c|}{\color{blue}{94.44}} & \multicolumn{1}{c|}{83.44} & \multicolumn{1}{c|}{93.82} & 89.18  \\ \hline
                 Shadows& \color{blue}{93.08} & \multicolumn{1}{c|}{88.55} & \multicolumn{1}{c|}{85.41} & \multicolumn{1}{c|}{87.42} & \multicolumn{1}{c|}{91.57} & \multicolumn{1}{c|}{89.43} & \multicolumn{1}{c|}{91.19} & \color{red}{93.96} & \multicolumn{1}{c|}{89.94} & \multicolumn{1}{c|}{89.43} & \multicolumn{1}{c|}{92.33} & \multicolumn{1}{c|}{\color{blue}{93.71}} & \multicolumn{1}{c|}{91.45} & \multicolumn{1}{c|}{\color{red}{94.34}} & 91.07  \\ \hline
                OA ($\uparrow$)& 74.27  & \multicolumn{1}{c|}{76.26} & \multicolumn{1}{c|}{80.77} & \multicolumn{1}{c|}{73.90} & \multicolumn{1}{c|}{74.89} & \multicolumn{1}{c|}{\color{blue}{82.97}} & \multicolumn{1}{c|}{82.67} & \color{red}{86.26} & \multicolumn{1}{c|}{73.61} & \multicolumn{1}{c|}{76.14} & \multicolumn{1}{c|}{76.30} & \multicolumn{1}{c|}{75.70} & \multicolumn{1}{c|}{75.46} & \multicolumn{1}{c|}{\color{blue}{79.69}} & \color{red}{87.31}  \\ \hline
                AA ($\uparrow$)& 79.01  & \multicolumn{1}{c|}{81.05} & \multicolumn{1}{c|}{80.67} & \multicolumn{1}{c|}{78.88} & \multicolumn{1}{c|}{79.63} & \multicolumn{1}{c|}{\color{blue}{83.37}} & \multicolumn{1}{c|}{82.15} & \color{red}{86.85}  & \multicolumn{1}{c|}{\color{blue}{82.78}} & \multicolumn{1}{c|}{79.97} & \multicolumn{1}{c|}{80.78} & \multicolumn{1}{c|}{80.30} & \multicolumn{1}{c|}{78.66} & \multicolumn{1}{c|}{79.67} & \color{red}{86.22}  \\ \hline
               $\kappa$ ($\uparrow$)  & 66.93  & \multicolumn{1}{c|}{69.62} & \multicolumn{1}{c|}{74.61} & \multicolumn{1}{c|}{66.91} & \multicolumn{1}{c|}{67.52} & \multicolumn{1}{c|}{\color{blue}{77.61}} & \multicolumn{1}{c|}{76.91} & \color{red}{81.96} & \multicolumn{1}{c|}{67.11} & \multicolumn{1}{c|}{68.93} & \multicolumn{1}{c|}{69.40} & \multicolumn{1}{c|}{68.89} & \multicolumn{1}{c|}{67.85} & \multicolumn{1}{c|}{\color{blue}{73.17}} & \color{red}{82.94}\\ \hline
\bottomrule
\end{tabular}}
\end{adjustbox}
\end{table*}

\begin{table*}[]
\centering
\caption{The OA, AA, $\kappa$ values (\%) of knowledge transfer from Pavia University to Indian Pines and from Houston to IndianPian}
\label{tab:comparison-huston-to-indianpian}
\begin{adjustbox}{width=\linewidth}
\setlength{\tabcolsep}{0.5mm}{
\begin{tabular}{c|c|ccccccc|ccccccc}
\hline
\bottomrule
\multirow{2}{*}{Class} & \multirow{2}{*}{Baseline} & \multicolumn{7}{c|}{Pavia University$\rightarrow$Indian Pines} & \multicolumn{7}{c}{Houston$\rightarrow$Indian Pines} \\ \cline{3-16} 
&                           & \multicolumn{1}{c|}{MTL} & \multicolumn{1}{c|}{UAN} & \multicolumn{1}{c|}{ONE} & \multicolumn{1}{c|}{FFL} & \multicolumn{1}{c|}{Adaptor} & \multicolumn{1}{c|}{Finetune} & Ours & \multicolumn{1}{c|}{MTL} & \multicolumn{1}{c|}{UAN} & \multicolumn{1}{c|}{ONE} & \multicolumn{1}{c|}{FFL} & \multicolumn{1}{c|}{Adaptor} & \multicolumn{1}{c|}{Finetune} & Ours \\ \hline

                 Alfalfa& \color{blue}{66.04} & \multicolumn{1}{c|}{59.10} & \multicolumn{1}{c|}{56.29} & \multicolumn{1}{c|}{61.42} & \multicolumn{1}{c|}{66.84} & \multicolumn{1}{c|}{\color{blue}{68.21}} & \multicolumn{1}{c|}{58.09} & \color{red}{73.99} & \multicolumn{1}{c|}{55.63} & \multicolumn{1}{c|}{48.77} & \multicolumn{1}{c|}{62.21} & \multicolumn{1}{c|}{58.53} & \multicolumn{1}{c|}{61.27} & \multicolumn{1}{c|}{64.09} & \color{red}{73.77} \\ \hline
                Corn-notill& \color{blue}{89.29}  & \multicolumn{1}{c|}{69.26} & \multicolumn{1}{c|}{73.09} & \multicolumn{1}{c|}{83.93} & \multicolumn{1}{c|}{80.10} & \multicolumn{1}{c|}{61.73} & \multicolumn{1}{c|}{76.53} & \color{red}{92.60}& \multicolumn{1}{c|}{68.11} & \multicolumn{1}{c|}{65.69} & \multicolumn{1}{c|}{\color{blue}{83.93}} & \multicolumn{1}{c|}{83.29} & \multicolumn{1}{c|}{83.16} & \multicolumn{1}{c|}{72.70} & 83.04 \\ \hline
                 Corn-mintill& \color{red}{99.46} & \multicolumn{1}{c|}{86.96} & \multicolumn{1}{c|}{89.67} & \multicolumn{1}{c|}{93.48} & \multicolumn{1}{c|}{\color{blue}{94.02}} & \multicolumn{1}{c|}{87.50} & \multicolumn{1}{c|}{92.39} & 92.93 & \multicolumn{1}{c|}{\color{blue}{95.65}} & \multicolumn{1}{c|}{89.13} & \multicolumn{1}{c|}{93.48} & \multicolumn{1}{c|}{95.11} & \multicolumn{1}{c|}{91.30} & \multicolumn{1}{c|}{90.22} & 94.02 \\ \hline
                 
                Corn& \color{red}{85.01} & \multicolumn{1}{c|}{83.67} & \multicolumn{1}{c|}{73.15} & \multicolumn{1}{c|}{84.34} & \multicolumn{1}{c|}{79.87} & \multicolumn{1}{c|}{78.75} & \multicolumn{1}{c|}{79.19} & \color{red}{85.01} & \multicolumn{1}{c|}{83.45} & \multicolumn{1}{c|}{78.97} & \multicolumn{1}{c|}{78.30} & \multicolumn{1}{c|}{82.77} & \multicolumn{1}{c|}{81.43} & \multicolumn{1}{c|}{82.33} & \color{blue}{84.79}  \\ \hline
                Grass-pasture& \color{blue}{91.39}& \multicolumn{1}{c|}{86.94} & \multicolumn{1}{c|}{78.05} & \multicolumn{1}{c|}{83.36} & \multicolumn{1}{c|}{88.67} & \multicolumn{1}{c|}{78.62} & \multicolumn{1}{c|}{78.77} &  \color{red}{91.97} & \multicolumn{1}{c|}{79.20} & \multicolumn{1}{c|}{66.86} & \multicolumn{1}{c|}{85.08} & \multicolumn{1}{c|}{79.34} & \multicolumn{1}{c|}{80.77} & \multicolumn{1}{c|}{75.61} & \color{blue}{91.39}  \\ \hline
                 Grass-trees& 93.62 & \multicolumn{1}{c|}{\color{red}{94.76}} & \multicolumn{1}{c|}{\color{red}{94.76}} & \multicolumn{1}{c|}{92.48} & \multicolumn{1}{c|}{93.62} & \multicolumn{1}{c|}{91.11} & \multicolumn{1}{c|}{92.26} & \color{blue}{93.85} & \multicolumn{1}{c|}{\color{red}{94.31}} & \multicolumn{1}{c|}{94.08} & \multicolumn{1}{c|}{92.71} & \multicolumn{1}{c|}{94.08} & \multicolumn{1}{c|}{93.17} & \multicolumn{1}{c|}{90.66} & \color{blue}{93.85}  \\ \hline 
                 Grass-pasture-mowed& \color{blue}{70.26} & \multicolumn{1}{c|}{60.46} & \multicolumn{1}{c|}{54.58} & \multicolumn{1}{c|}{49.89} & \multicolumn{1}{c|}{68.52} & \multicolumn{1}{c|}{62.96} & \multicolumn{1}{c|}{62.31} & \color{red}{70.92} & \multicolumn{1}{c|}{63.40} & \multicolumn{1}{c|}{66.56} & \multicolumn{1}{c|}{66.56} & \multicolumn{1}{c|}{61.33} & \multicolumn{1}{c|}{69.93} & \multicolumn{1}{c|}{55.12} & \color{red}{72.77}  \\ \hline
                 Hay-windrowed& 64.19  & \multicolumn{1}{c|}{58.60} & \multicolumn{1}{c|}{33.46} & \multicolumn{1}{c|}{57.73} & \multicolumn{1}{c|}{64.31} & \multicolumn{1}{c|}{\color{blue}{68.78}} & \multicolumn{1}{c|}{65.34} & \color{red}{72.50}  & \multicolumn{1}{c|}{50.95} & \multicolumn{1}{c|}{41.52} & \multicolumn{1}{c|}{56.99} & \multicolumn{1}{c|}{60.26} & \multicolumn{1}{c|}{\color{blue}{66.67}} & \multicolumn{1}{c|}{65.34} & \color{red}{74.65} \\ \hline
                 Oats& \color{red}{76.06} & \multicolumn{1}{c|}{70.39} & \multicolumn{1}{c|}{55.67} & \multicolumn{1}{c|}{68.44} & \multicolumn{1}{c|}{64.89} & \multicolumn{1}{c|}{67.02} & \multicolumn{1}{c|}{66.31} & \color{blue}{73.94} & \multicolumn{1}{c|}{66.31} & \multicolumn{1}{c|}{71.45} & \multicolumn{1}{c|}{68.44} & \multicolumn{1}{c|}{\color{blue}{74.65}} & \multicolumn{1}{c|}{47.70} & \multicolumn{1}{c|}{59.22} & 66.13 \\ \hline  
                  Soybean-notill& 95.68& \multicolumn{1}{c|}{\color{red}{98.15}} & \multicolumn{1}{c|}{96.91} & \multicolumn{1}{c|}{\color{blue}{97.53}} & \multicolumn{1}{c|}{95.06} & \multicolumn{1}{c|}{96.91} & \multicolumn{1}{c|}{95.06} & 96.30 & \multicolumn{1}{c|}{94.44} & \multicolumn{1}{c|}{91.98} & \multicolumn{1}{c|}{95.68} & \multicolumn{1}{c|}{\color{blue}{98.77}} & \multicolumn{1}{c|}{88.27} & \multicolumn{1}{c|}{97.53} & \color{red}{99.38} \\ \hline
                  Soybean-mintill& \color{blue}{93.89}& \multicolumn{1}{c|}{\color{red}{94.21}} & \multicolumn{1}{c|}{92.12} & \multicolumn{1}{c|}{81.91} & \multicolumn{1}{c|}{90.19} & \multicolumn{1}{c|}{84.81} & \multicolumn{1}{c|}{87.86} & 91.24 & \multicolumn{1}{c|}{81.83} & \multicolumn{1}{c|}{83.04} & \multicolumn{1}{c|}{81.03} & \multicolumn{1}{c|}{\color{blue}{93.17}} & \multicolumn{1}{c|}{83.52} & \multicolumn{1}{c|}{91.80} & 92.36  \\ \hline
                 Soybean-clean& \color{red}{87.88} & \multicolumn{1}{c|}{64.85} & \multicolumn{1}{c|}{72.12} & \multicolumn{1}{c|}{80.91} & \multicolumn{1}{c|}{74.24} & \multicolumn{1}{c|}{87.27} & \multicolumn{1}{c|}{78.79} & \color{red}{87.88}& \multicolumn{1}{c|}{77.58} & \multicolumn{1}{c|}{77.58} & \multicolumn{1}{c|}{78.48} & \multicolumn{1}{c|}{81.52} & \multicolumn{1}{c|}{83.94} & \multicolumn{1}{c|}{72.42} & \color{red}{88.18} \\ \hline
                 Wheat & 95.56 & \multicolumn{1}{c|}{\color{red}{100.00 }} & \multicolumn{1}{c|}{\color{red}{100.00 }} & \multicolumn{1}{c|}{\color{red}{100.00 }} & \multicolumn{1}{c|}{\color{red}{100.00 }} & \multicolumn{1}{c|}{\color{red}{100.00 }} & \multicolumn{1}{c|}{\color{red}{100.00 }} & \color{blue}{97.78} & \multicolumn{1}{c|}{\color{red}{100.00 }} & \multicolumn{1}{c|}{\color{blue}{97.78}} & \multicolumn{1}{c|}{\color{red}{100.00 }} & \multicolumn{1}{c|}{\color{red}{100.00 }} & \multicolumn{1}{c|}{93.33} & \multicolumn{1}{c|}{\color{red}{100.00 }} & \color{red}{100.00 } \\ \hline
                 Woods & 89.74& \multicolumn{1}{c|}{\color{blue}{97.44}} & \multicolumn{1}{c|}{\color{blue}{97.44}} & \multicolumn{1}{c|}{94.87} & \multicolumn{1}{c|}{\color{red}{100.00 }} & \multicolumn{1}{c|}{\color{red}{100.00 }} & \multicolumn{1}{c|}{\color{blue}{97.44}} & \color{red}{100.00 } & \multicolumn{1}{c|}{\color{red}{100.00 }} & \multicolumn{1}{c|}{\color{blue}{97.44}} & \multicolumn{1}{c|}{\color{red}{100.00 }} & \multicolumn{1}{c|}{\color{red}{100.00 }} & \multicolumn{1}{c|}{\color{red}{100.00 }} & \multicolumn{1}{c|}{89.74} & \color{red}{100.00 } \\ \hline
                Buildings-Grass-Trees-Drives&\color{red}{100.00 }   & \multicolumn{1}{c|}{\color{red}{100.00 }} & \multicolumn{1}{c|}{\color{red}{100.00 }} & \multicolumn{1}{c|}{\color{red}{100.00 }} & \multicolumn{1}{c|}{\color{red}{100.00 }} & \multicolumn{1}{c|}{\color{red}{100.00 }} & \multicolumn{1}{c|}{\color{red}{100.00 }} & \color{red}{100.00 } & \multicolumn{1}{c|}{\color{red}{100.00 }} & \multicolumn{1}{c|}{\color{red}{100.00 }} & \multicolumn{1}{c|}{\color{red}{100.00 }} & \multicolumn{1}{c|}{\color{red}{100.00 }} & \multicolumn{1}{c|}{\color{red}{100.00 }} & \multicolumn{1}{c|}{\color{red}{100.00 }} &\color{red}{100.00 }  \\ \hline
                 Stone-Steel-Towers&\color{red}{100.00 }  & \multicolumn{1}{c|}{\color{red}{100.00 }} & \multicolumn{1}{c|}{\color{red}{100.00 }} & \multicolumn{1}{c|}{\color{red}{100.00 }} & \multicolumn{1}{c|}{\color{red}{100.00 }} & \multicolumn{1}{c|}{\color{red}{100.00 }} & \multicolumn{1}{c|}{\color{red}{100.00 }} & \color{red}{100.00 } & \multicolumn{1}{c|}{\color{red}{100.00 }} & \multicolumn{1}{c|}{\color{red}{100.00 }} & \multicolumn{1}{c|}{\color{red}{100.00 }} & \multicolumn{1}{c|}{\color{red}{100.00 }} & \multicolumn{1}{c|}{\color{red}{100.00 }} & \multicolumn{1}{c|}{\color{red}{100.00 }} &\color{red}{100.00 }  \\ \hline
                OA ($\uparrow$) & \color{blue}{78.15} & \multicolumn{1}{c|}{71.66} & \multicolumn{1}{c|}{62.74} & \multicolumn{1}{c|}{70.58} & \multicolumn{1}{c|}{75.31} & \multicolumn{1}{c|}{73.50} & \multicolumn{1}{c|}{72.53} & \color{red}{81.21} & \multicolumn{1}{c|}{67.56} & \multicolumn{1}{c|}{63.49} & \multicolumn{1}{c|}{71.78} & \multicolumn{1}{c|}{73.45} & \multicolumn{1}{c|}{73.24} & \multicolumn{1}{c|}{72.08} & \color{red}{80.86}   \\ \hline
                AA ($\uparrow$)& \color{blue}{87.38}  & \multicolumn{1}{c|}{ 82.80} & \multicolumn{1}{c|}{79.21} & \multicolumn{1}{c|}{83.14} & \multicolumn{1}{c|}{85.02} & \multicolumn{1}{c|}{83.36} & \multicolumn{1}{c|}{83.15} & \color{red}{88.77} & \multicolumn{1}{c|}{81.93} & \multicolumn{1}{c|}{79.43} & \multicolumn{1}{c|}{84.08} & \multicolumn{1}{c|}{85.17} & \multicolumn{1}{c|}{82.78} & \multicolumn{1}{c|}{81.67} & \color{red}{88.40} \\ \hline
                $\kappa$ ($\uparrow$) & \color{blue}{75.30} & \multicolumn{1}{c|}{68.12} & \multicolumn{1}{c|}{58.62} & \multicolumn{1}{c|}{66.90} & \multicolumn{1}{c|}{72.06} & \multicolumn{1}{c|}{69.84} & \multicolumn{1}{c|}{68.92} & \color{red}{78.67} & \multicolumn{1}{c|}{63.79} & \multicolumn{1}{c|}{59.41} & \multicolumn{1}{c|}{68.28} & \multicolumn{1}{c|}{70.17} & \multicolumn{1}{c|}{69.65} & \multicolumn{1}{c|}{68.23} & \color{red}{78.22}  \\ \hline
\bottomrule
\end{tabular}}
\end{adjustbox}
\end{table*}


In Tables \ref{tab:comparison-indianpian-to-huston} and \ref{tab:comparison-huston-to-pavia}, our method achieves the best performance across four different cross-transfer settings. Additionally, most knowledge transfer methods outperform the baseline. Specifically, when using Houston as the target scene, our transfer from Pavia University to Houston yields gains of 4.68\%, 3.98\%, and 5.07\% over the second-best method for OA, AA, and $\kappa$, respectively. Similarly, the transfer from Indian Pines to Houston results in improvements of 1.09\%, 0.25\%, and 1.19\% over the second-best method for OA, AA, and $\kappa$, respectively. The results indicate that transferring knowledge from Pavia University to Houston using our method yields better performance than transferring from Indian Pines to Houston, suggesting that the size of the source dataset impacts performance. The larger Pavia University dataset—approximately four times the size of Indian Pines—enables the learning of more robust representations, reduces overfitting to specific instances, and improves generalization to new, unseen data.
When Pavia University is the target scene, the results are similar to those for Houston. Our method shows significant performance improvements when transferring from Indian Pines to Pavia University, with gains of 3.29\%, 3.48\%, and 4.35\% over the second-best method for OA, AA, and $\kappa$, respectively. When transferring from Houston to Pavia University, the improvements are 7.62\%, 3.44\%, and 9.77\% for OA, AA, and $\kappa$, respectively. Since the size of Houston is 1.5 times larger than the Indian Pines dataset, our method performs better when transferring from Houston to Pavia University than from Indian Pines to Pavia University.

In Table V, most knowledge transfer methods perform worse than the baseline. 
This is mainly because the Indian Pines dataset has a broader variety of categories than the other two datasets.  Specifically, 56\% of its categories are fine-grained classes, such as corn-notill, corn-mintill, corn, grass-pasture, grass-trees, grass-pasture-mowed, soybean-notill, soybean-mintill, and soybean-clean, and do not have direct semantic counterparts in the Houston or Pavia University datasets.
Because of this, when Indian Pines dataset serves as target dataset, it is difficult to improve the performance since the source scene is not closely related to the target scene \cite{rosenstein2005transfer,weiss2016survey,day2017survey}. 
Enforcing alignment between source and target distributions and label spaces can result in over-adaptation, where unrelated features are forced into a shared latent feature space, ultimately hindering target performance \cite{bao2023survey}. However, our method still achieves the best performance. 
We attribute this to the ASC and CKSP components, which align the spectral distributions and semantic space while simultaneously determining the significance of each source sample in the process of knowledge sharing, thereby mitigating the impact of negative transfer.

\subsection{Comparison of Model Performance with Varied Target Training Samples}
We conducted experiments with 10, 50, and full target training samples per category. Under the same conditions, we compared the performance of our knowledge transfer method with the baseline, as shown in Tables VI, VII, and VIII.

Tables VI and VIII demonstrate that our method consistently outperforms the baseline under the same target training sample conditions. This indicates that our approach more effectively integrates the weakly related auxiliary dataset (source scene) to enhance target performance. 
Moreover, when the target scene has very limited data samples, our method leverage source scene information to achieve results comparable to the baseline trained on the full target set.
Thus, with very limited target samples, our method makes effective use of the source data to produce promising outcomes.
As the number of training samples per category grows, however, the performance gains from our knowledge transfer approach diminish compared to the baseline, because the need for auxiliary training decreases when more target data are available—making knowledge transfer less necessary \cite{rosenstein2005transfer}.

Similarly, Table VII shows that our method improves target-scene performance under varying training-sample conditions. However, the higher diversity and more fine-grained categories in the Indian Pines Pines dataset compared to the Houston and Pavia University scenes, present an inherent limitation on effective knowledge transfer when Indian Pines has limited training data. Notably, with 10 samples per category (approximately 10\% of the Indian Pines training data), our approach shows a slight performance drop compared to the baseline trained on the full training set.

\begin{table*}[]
\centering
\caption{Results for Pavia University with 10, 50, and full training target samples per category. H$\rightarrow$P and I$\rightarrow$P represent our method leveraging knowledge from the Houston and Indian Pines datasets, respectively. Additionally, P represents the results of the baseline model.
}
\begin{tabular}{c|ccc|ccc|ccc}
\hline
\bottomrule
\multirow{2}{*}{Performance (\%)} & \multicolumn{3}{c|}{10}                                       & \multicolumn{3}{c|}{50}                                       & \multicolumn{3}{c}{Full training set}                            \\ \cline{2-10} 
                             & \multicolumn{1}{c|}{H$\rightarrow$P}   & \multicolumn{1}{c|}{I$\rightarrow$P}   & P      & \multicolumn{1}{c|}{H$\rightarrow$P}   & \multicolumn{1}{c|}{I$\rightarrow$P}   & P      & \multicolumn{1}{c|}{H$\rightarrow$P}   & \multicolumn{1}{c|}{I$\rightarrow$P} & P      \\ \hline
OA ($\uparrow$)                         & \multicolumn{1}{c|}{87.31} & \multicolumn{1}{c|}{86.26} & 74.27 & \multicolumn{1}{c|}{87.41} & \multicolumn{1}{c|}{86.88} & 85.73 & \multicolumn{1}{c|}{87.99} & \multicolumn{1}{c|}{86.95}     & 86.24  \\ \hline
AA ($\uparrow$)                         & \multicolumn{1}{c|}{86.22} & \multicolumn{1}{c|}{86.85} & 79.01 & \multicolumn{1}{c|}{86.90} & \multicolumn{1}{c|}{84.74} & 85.55 & \multicolumn{1}{c|}{85.63} & \multicolumn{1}{c|}{84.58}     & 85.20 \\ \hline
$\kappa$ ($\uparrow$)      & \multicolumn{1}{c|}{82.94} & \multicolumn{1}{c|}{81.96} & 66.93 & \multicolumn{1}{c|}{83.32} & \multicolumn{1}{c|}{82.33} & 80.91 & \multicolumn{1}{c|}{83.89} & \multicolumn{1}{c|}{82.32}     & 81.64 \\ \hline
\bottomrule
\end{tabular}
\end{table*}


\begin{table*}[]
\centering
\caption{Results for Indian Pines with 10, 50, and full training target samples per category. H$\rightarrow$I and P$\rightarrow$I represent our method leveraging knowledge from the Houston and Pavia University datasets, respectively. Additionally, I represents the results of the baseline model.}
\begin{tabular}{c|ccc|ccc|ccc}
\hline
\bottomrule
\multirow{2}{*}{Performance (\%)} & \multicolumn{3}{c|}{10}                                       & \multicolumn{3}{c|}{50}                                       & \multicolumn{3}{c}{Full training set}                              \\ \cline{2-10} 
                             & \multicolumn{1}{c|}{H$\rightarrow$I}   & \multicolumn{1}{c|}{P$\rightarrow$I}   & I      & \multicolumn{1}{c|}{H$\rightarrow$I}   & \multicolumn{1}{c|}{P$\rightarrow$I}   & I      & \multicolumn{1}{c|}{H$\rightarrow$I}   & \multicolumn{1}{c|}{P$\rightarrow$I}   & I      \\ \hline
OA ($\uparrow$)                        & \multicolumn{1}{c|}{80.86} & \multicolumn{1}{c|}{81.21} & 78.15 & \multicolumn{1}{c|}{83.50} & \multicolumn{1}{c|}{84.58} & 83.39 & \multicolumn{1}{c|}{85.51} & \multicolumn{1}{c|}{86.02} & 84.22 \\ \hline
AA ($\uparrow$)                          & \multicolumn{1}{c|}{88.40} & \multicolumn{1}{c|}{88.77} & 87.38 & \multicolumn{1}{c|}{90.33} & \multicolumn{1}{c|}{90.38} & 89.59 & \multicolumn{1}{c|}{91.15} & \multicolumn{1}{c|}{91.48} & 92.67 \\ \hline
$\kappa$ ($\uparrow$)                      & \multicolumn{1}{c|}{78.22} & \multicolumn{1}{c|}{78.67} & 75.30 & \multicolumn{1}{c|}{81.28} & \multicolumn{1}{c|}{82.46} & 80.89 & \multicolumn{1}{c|}{83.53} & \multicolumn{1}{c|}{84.08} & 82.15 \\ \hline
\bottomrule
\end{tabular}
\end{table*}


\begin{table*}[]
\centering
\caption{Results for Houston with 10, 50, and full training target samples per category. I$\rightarrow$H and P$\rightarrow$H represent our method leveraging knowledge from the Indian Pines and Pavia University datasets, respectively. Additionally, H represents the results of the baseline model.}
\begin{tabular}{c|ccc|ccc|ccc}
\hline
\bottomrule
\multirow{2}{*}{Performance (\%)} & \multicolumn{3}{c|}{10}                                       & \multicolumn{3}{c|}{50}                                       & \multicolumn{3}{c}{Full training set}                          \\ \cline{2-10} 
                             & \multicolumn{1}{c|}{I$\rightarrow$H}   & \multicolumn{1}{c|}{P$\rightarrow$H}   & H      & \multicolumn{1}{c|}{I$\rightarrow$H}   & \multicolumn{1}{c|}{P$\rightarrow$H}   & H      & \multicolumn{1}{c|}{I$\rightarrow$H} & \multicolumn{1}{c|}{P$\rightarrow$H} & H      \\ \hline
OA                           & \multicolumn{1}{c|}{82.82} & \multicolumn{1}{c|}{84.64} & 79.24 & \multicolumn{1}{c|}{83.91} & \multicolumn{1}{c|}{84.71} & 83.30 & \multicolumn{1}{c|}{85.75}     & \multicolumn{1}{c|}{86.06}     & 85.44 \\ \hline
AA                           & \multicolumn{1}{c|}{82.46} & \multicolumn{1}{c|}{84.00} & 78.53 & \multicolumn{1}{c|}{84.18} & \multicolumn{1}{c|}{84.17} & 84.64 & \multicolumn{1}{c|}{84.40}     & \multicolumn{1}{c|}{86.84}     & 86.04 \\ \hline
kappa                        & \multicolumn{1}{c|}{81.37} & \multicolumn{1}{c|}{83.33} & 77.52 & \multicolumn{1}{c|}{82.54} & \multicolumn{1}{c|}{83.40} & 81.86 & \multicolumn{1}{c|}{82.42}     & \multicolumn{1}{c|}{84.86}     & 84.19 \\ \hline
\bottomrule
\end{tabular}
\end{table*}

\subsection{Ablation Study}

\textbf{Impact of various components:}
Tables \ref{ablation_H_P} and \ref{ablation_I_H} present the results of an ablation study demonstrating the individual effectiveness of key  components in our method, including ACS, CKSP, and CII (comprising CE and DI). 

As shown in Tables \ref{ablation_H_P} and \ref{ablation_I_H}, introducing ACS leads to performance gains of 2.50\% for OA, 0.30\% for AA, and 2.71\% for $\kappa$ when transferring from Houston to Pavia University, with 2.95\% for OA, 2.06\% for AA, and 3.15\% for $\kappa$ from Indian Pines to Houston scenes, validating ACS’s role in aligning source and target domains.
The CKSP component increases performance by 2.70\% for OA, 1.78\% for AA, and 3.43\% for $\kappa$ from Houston to Pavia University, and by 1.05\% for OA, 1.12\% for AA, and 1.16\% for $\kappa$ from Indian Pines to Houston, indicating CKSP’s ability to mitigate mismatched semantics.
Introducing CE alone yields only a minor improvement, suggesting that shared and complementary information are difficult to fuse without the DI module. By contrast, DI provides the most substantial boosts, reaching 6.71\% for OA, 5.67\% for AA, and 8.61\% for $\kappa$ from Houston to Pavia University scenes, and 4.46\% for OA, 4.02\% for AA, and 4.84\% for $\kappa$ from Indian Pines to Houston scenes. These findings underscore the importance of complementary information and highlight that integrating shared and complementary information is crucial for maintaining the completeness of target data.

\begin{table}[]
\centering
\caption{The ablation study results of knowledge transfer from Houston to Pavia University.}
\begin{tabular}{cccc|ccc}
\hline
\bottomrule
\multicolumn{4}{c|}{Components}                                                                                  & \multicolumn{3}{c}{Performance (\%)}                                    \\ \hline
\multicolumn{1}{c}{ACS}        & \multicolumn{1}{c}{CKSP}       & \multicolumn{1}{c}{CE}          & DI         & \multicolumn{1}{c|}{OA}     & \multicolumn{1}{c|}{AA}     & $\kappa$  \\ \hline
\multicolumn{1}{c}{\ding{55}} & \multicolumn{1}{c}{\ding{55}} & \multicolumn{1}{c}{\ding{55}} & \ding{55} & \multicolumn{1}{c|}{73.98} & \multicolumn{1}{c|}{78.46} & 66.49 \\ \hline
\multicolumn{1}{c}{\ding{51}} & \multicolumn{1}{c}{\ding{55}} & \multicolumn{1}{c}{\ding{55}} & \ding{55} & \multicolumn{1}{c|}{76.48} & \multicolumn{1}{c|}{78.76} & 69.20 \\ \hline
\multicolumn{1}{c}{\ding{51}} & \multicolumn{1}{c}{\ding{51}} & \multicolumn{1}{c}{\ding{55}} & \ding{55} & \multicolumn{1}{c|}{79.18} & \multicolumn{1}{c|}{80.54} & 72.63 \\ \hline
\multicolumn{1}{c}{\ding{51}} & \multicolumn{1}{c}{\ding{51}} & \multicolumn{1}{c}{\ding{51}} & \ding{55} & \multicolumn{1}{c|}{80.60} & \multicolumn{1}{c|}{80.55} & 74.33 \\ \hline
\multicolumn{1}{c}{\ding{51}} & \multicolumn{1}{c}{\ding{51}} & \multicolumn{1}{c}{\ding{51}} & \ding{51} & \multicolumn{1}{c|}{\color{red}{87.31}} & \multicolumn{1}{c|}{\color{red}{86.22}} & \color{red}{82.94} \\ \hline
\bottomrule
\end{tabular}
\label{ablation_H_P}
\end{table}

\begin{table}[]
\centering
\caption{The ablation study results of knowledge transfer from Indian Pines to Houston.}
\begin{tabular}{cccc|ccc}
\hline
\bottomrule
\multicolumn{4}{c|}{Components}                                                                                  & \multicolumn{3}{c}{Performance (\%)}                                    \\ \hline
\multicolumn{1}{c}{ACS}        & \multicolumn{1}{c}{CKSP}       & \multicolumn{1}{c}{CE}          & DI         & \multicolumn{1}{c|}{OA}     & \multicolumn{1}{c|}{AA}     & $\kappa$  \\ \hline
\multicolumn{1}{c}{\ding{55}} & \multicolumn{1}{c}{\ding{55}} & \multicolumn{1}{c}{\ding{55}} & \ding{55} & \multicolumn{1}{c|}{74.72} & \multicolumn{1}{c|}{74.81} & 72.59 \\ \hline
\multicolumn{1}{c}{\ding{51}} & \multicolumn{1}{c}{\ding{55}} & \multicolumn{1}{c}{\ding{55}} & \ding{55} & \multicolumn{1}{c|}{77.67} & \multicolumn{1}{c|}{76.87} & 75.74 \\ \hline
\multicolumn{1}{c}{\ding{51}} & \multicolumn{1}{c}{\ding{51}} & \multicolumn{1}{c}{\ding{55}} & \ding{55} & \multicolumn{1}{c|}{78.72} & \multicolumn{1}{c|}{77.99} & 76.90 \\ \hline
\multicolumn{1}{c}{\ding{51}} & \multicolumn{1}{c}{\ding{51}} & \multicolumn{1}{c}{\ding{51}} & \ding{55} & \multicolumn{1}{c|}{78.36} & \multicolumn{1}{c|}{78.44} & 76.53 \\ \hline
\multicolumn{1}{c}{\ding{51}} & \multicolumn{1}{c}{\ding{51}} & \multicolumn{1}{c}{\ding{51}} & \ding{51} & \multicolumn{1}{c|}{\color{red}{82.82}} & \multicolumn{1}{c|}{\color{red}{82.46}} & \color{red}{81.37} \\ \hline
\bottomrule
\end{tabular}
\label{ablation_I_H}
\end{table}

\textbf{Impact of Temperature:} 
In Tables. \ref{ablation_H_P} and \ref{ablation_I_H}, 
the DI component provides the most significant performance gains.
In Eq. \eqref{kt_1} and Eq. \eqref{kt_2}, $\tau$ controls the softness under teacher supervision in DI component. In Eq. \eqref{kt_1}, $\tau$ is fixed at 1.0. To investigate the sensitivity of $\tau$ in Eq. \eqref{kt_2}, we perform experiments using our method on the Houston to Pavia University and Indian Pines to Houston.

\begin{figure*}[!t]
\centering
\subfloat[]{\includegraphics[width=1.65in]{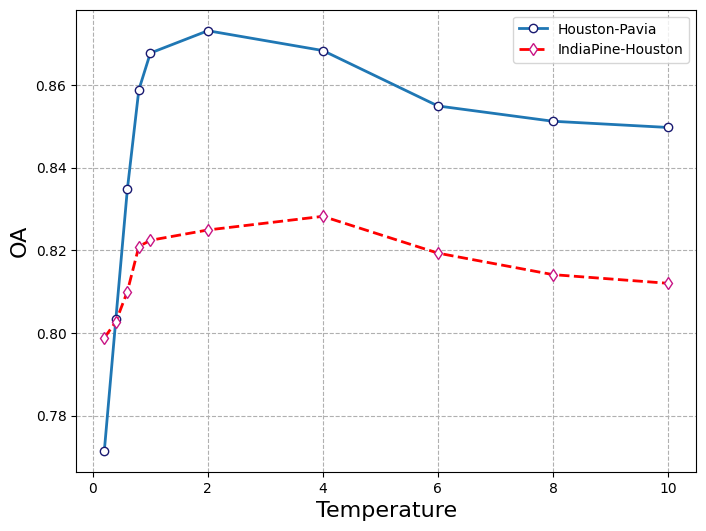}}
\hfil
\subfloat[]{\includegraphics[width=2in]{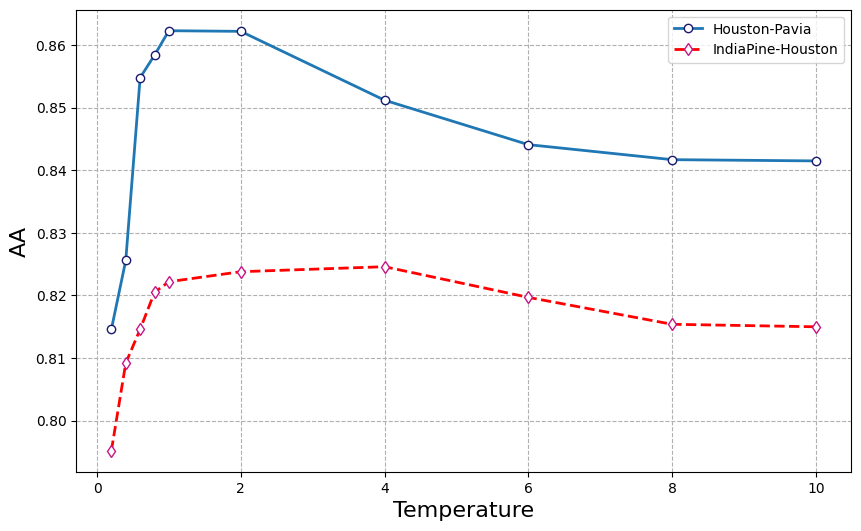}%
\label{second_case}}
\hfil
\subfloat[]{\includegraphics[width=2in]{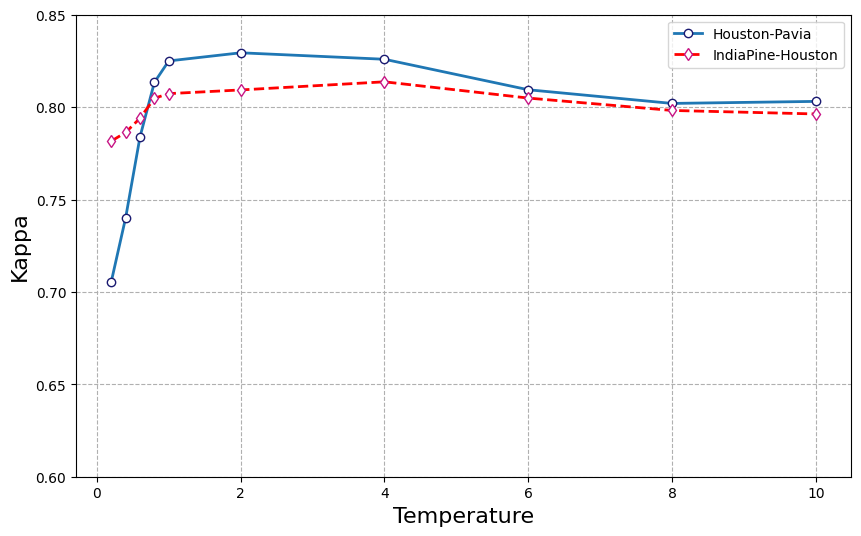}%
\label{third_case}}
\hfil
\caption{Performance of knowledge transfer with varying temperature: (a) from Houston to Pavia University, and (b) from Indian Pines to Houston.}
\label{temperature}
\end{figure*}

As $\tau$ increases, the output of teacher becomes smoother, resembling the uniform distribution of label smoothing. As shown in Fig. \ref{temperature}, our method achieves higher gains when $\tau >1$ whereas a small $\tau$ limits the benefits of distillation. Lower temperatures sharpen the probability distribution of the complementary module, leading the distillation process to rely more heavily on maximal logits. This causes the loss of semantic similarities across different classes, making it difficult for the student model to learn the semantic similarities within the complementary module and to better complete the information learned from the shared portion. In contrast, a higher temperature softens the probability distribution, allowing the distillation process to focus more on the richer, inter-class information. This enables the student model to learn additional class similarities from the complementary module and to better calibrate the similarities between the complementary and shared portions.

\section{Conclusion}
We highlight three distinct challenges: spectral differences among HSI sensors, the lack of direct category correspondence across different datasets, and the incomplete information in unmatched portions when transferring knowledge from HSI datasets under real-world conditions. 
Our approach, CKI, addresses the spectral discrepancies by ASC in a domain-agnostic space. Additionally, CKI identifies mismatched semantics through CKSP, leveraging the Source Similarity Mechanism. Furthermore,  we introduce CII to resolve target-private information utilization deficiencies in shared knowledge through CE and DI modules. As a result, CKI demonstrates stable performance across diverse knowledge transfer settings and achieves state-of-the-art results.

\bibliographystyle{IEEEtran}
\bibliography{IEEEexample}
\clearpage
\newpage
\appendices 
\section*{The structure of IFSS Transformer} \label{sec:apdx-ifss}
The IFSS Transformer $G$ extracts spatial feature through a Spatial Transformer block and spectral feature through a Spectral Transformer block \cite{scheibenreif2023masked}, then combined these two different types of features through an Interaction Fusion block \cite{huo2024center}.
\subsection*{Spatial and Spectral Transformer}
The spatial and spectral transformers divide an input image  $\mathbf{x} \in \mathbb{R}^{H \times W \times C} $ into non-overlapping patches $ \mathbf{p} \in \mathbb{R}^{N \times (p^h \times p^w \times p^c)}$, where $H$, $W$, and $C$ are the height, width, and number of channels of the input image, respectively, and $p^h$, $p^w$ and $p^c$ represent the height, width, and number of channels of each patch. Here, $N = \left(\frac{H}{p^h}\right) \times \left(\frac{W}{p^w}\right) \times \left(\frac{C}{p^c}\right)$ denotes the total number of patches.

Each patch in the Spatial Transformer is then linearly embedded into a dimension $D$, and positional encodings are added to retain spatial information. In contrast, the Spectral Transformer employs multiple distinct spectral wavelength intervals for each spatial patch. A block-wise spectral embedding utilizes a distinct embedding for each $\frac{C}{p^c}$ spectral block.

The resulting spatial and spectral features are then computed by multi-head self-attention (MSA) and feed-forward network (FFN) layers \cite{vaswani2017attention}, both with layer normalization (LN) \cite{lei2016layer}:

\begin{align}
&\mathbf{y}^l_i = \text{MSA}(\text{LN}(\mathbf{z}^{l}_i)) + \mathbf{z}^{l}_i,\\
&\mathbf{z}^{l+1}_i = \text{FFN}(\text{LN}(\mathbf{y}^l_i)) + \mathbf{y}^l_i.
\end{align}
where $l$ denotes the layer number and $i \in \{spectral, spatial\}$ represents the spectral and spatial features, respectively.

\subsection*{Interaction Fusion}
The Interaction Fusion contains two components including fusion matrix and feature fusion.
The fusion matrix $z^{m}$ is defined by four views $z_{ij}$, for $i\textsc{,}j\in\{1,2\}$, where $1,2$ correspond to spectral and spatial features respectively. This matrix contains two intra-modality features ($z_{11}$ and $z_{22}$) computed by self-attention, and two inter-modality features ($z_{12}$ and $z_{21}$) computed by cross-attention. 
\begin{equation}\label{fusion-matrix}
\begin{aligned}
z^{m} = \begin{bmatrix} z_{11} & z_{12}\\ z_{21} & z_{22}\end{bmatrix} 
\end{aligned}
\end{equation}
The feature fusion layer $F_{\text{comp}}$ contains Layer Norm (LN), Convolutional (Conv), GELU, and residual layers, effectively integrating the information from different views into a combined feature $\hat{z}$.
\begin{equation}\label{FeatureFusion}
\begin{aligned}
\hat{z} = F_{\text{comp}}(z^{m})= \text{Conv}(\text{GELU}(\text{Conv}(\text{Conv}(\text{LN}(z^{m})))) + z^{m})
\end{aligned}
\end{equation}
Finally, the resulting feature $z$ of the IFSS transformer is computed by the FFN and the residual layer:
\begin{equation}\label{cross-attention}
\begin{aligned}
z = \hat{z} + \text{FFN}(\text{LN}(\hat{z}))
\end{aligned}
\end{equation}


\vfill

\end{document}